\documentclass[10pt,twocolumn,letterpaper]{article}

\usepackage[pagenumbers]{iccv} % To force page numbers, e.g. for an arXiv version

\usepackage{stfloats}
\usepackage{multirow} 
\usepackage[table,xcdraw]{xcolor}

\usepackage{cuted}

\definecolor{iccvblue}{rgb}{0.21,0.49,0.74}
\usepackage[pagebackref,breaklinks,colorlinks,allcolors=iccvblue]{hyperref}

\def\confName{ICCV}
\def\confYear{2025}

\definecolor{deepPink}{RGB}{255, 20, 147}

\title{AffordDexGrasp: Open-set Language-guided Dexterous Grasp \\with Generalizable-Instructive Affordance}

\author{
    \textbf{Yi-Lin Wei}\footnotemark[1] \textsuperscript{1},
    \quad \textbf{Mu Lin}\footnotemark[1]  \textsuperscript{1},
    \quad \textbf{Yuhao Lin} \textsuperscript{1},
    \quad \textbf{Jian-Jian Jiang} \textsuperscript{1}, \\
    \quad \textbf{Xiao-Ming Wu} \textsuperscript{1},
    \quad \textbf{Ling-An Zeng} \textsuperscript{1},
    \quad \textbf{Wei-Shi Zheng}\footnotemark[2] \textsuperscript{1,2} \\ 
    \textsuperscript{1} School of Computer Science and Engineering,  Sun Yat-sen University,  China \\
    \textsuperscript{2} Key Laboratory of Machine Intelligence and Advanced Computing, 
    Ministry of Education, China \\
    \footnotesize 
    \tt{
    \{weiylin5, linm67, linyh96, jiangjj35, wuxm65, zenglan3\}@mail2.sysu.edu.cn \quad wszheng@ieee.org    
    } \\
    \href{https://isee-laboratory.github.io/AffordDexGrasp/}{\textcolor{deepPink}{https://isee-laboratory.github.io/AffordDexGrasp/}}
}

\begin{document}
\maketitle
\captionsetup{hypcap=false} % 关闭 hypcap 以避免警告
\begin{strip}\centering
\vspace{-1.4cm}
\includegraphics[width=1\linewidth]{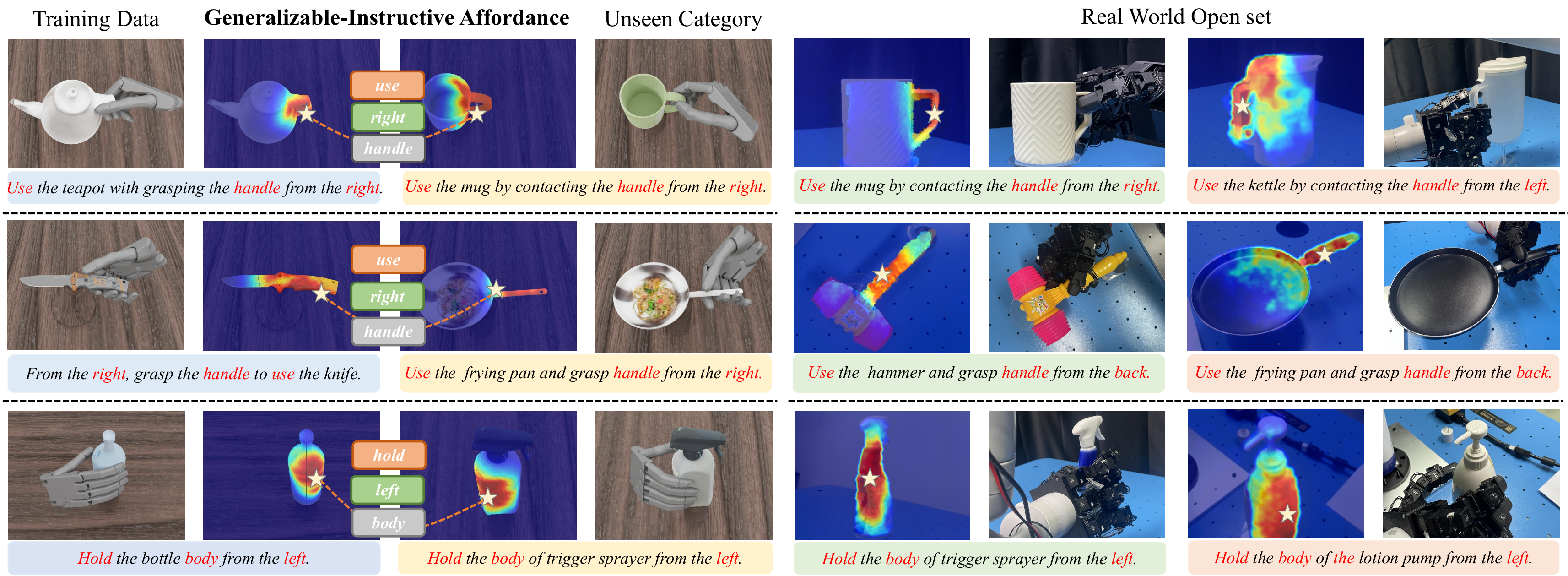}
\vspace{-0.5cm}
\captionof{figure}{
Open-set Language-guided Dexterous Grasp. Our framework bridges the gap between language and grasp actions through Generalizable-Instructive Affordance, which enables cross-category generalization via category-agnostic cues and graspable local structure. Remarkably, our framework demonstrates strong generalization without requiring extra real training data in real-world experiments.
}
\label{first_figure} % 修改拼写错误
\vspace{-.2cm}
\end{strip}

\footnotetext[1]{Equal contribution.}
\footnotetext[2]{Corresponding author.}

\begin{abstract}
Language-guided robot dexterous generation enables robots to grasp and manipulate objects based on human commands. However, previous data-driven methods are hard to understand intention and execute grasping with unseen categories in the open set. In this work, we explore a new task, Open-set Language-guided Dexterous Grasp, and find that the main challenge is the huge gap between high-level human language semantics and low-level robot actions. To solve this problem, we propose an Affordance Dexterous Grasp (AffordDexGrasp) framework, with the insight of bridging the gap with a new generalizable-instructive affordance representation. This affordance can generalize to unseen categories by leveraging the object's local structure and category-agnostic semantic attributes, thereby effectively guiding dexterous grasp generation. Built upon the affordance, our framework introduces Affordance Flow Matching (AFM) for affordance generation with language as input, and Grasp Flow Matching (GFM) for generating dexterous grasp with affordance as input. To evaluate our framework, we build an open-set table-top language-guided dexterous grasp dataset. Extensive experiments in the simulation and real worlds show that our framework surpasses all previous methods in open-set generalization.
\end{abstract}

\section{Introduction}
\label{sec:intro}

Achieving generalizable robot dexterous grasping is an important goal in the fields of robotics and computer vision, with exciting potential applications in human-robot interaction and robot manipulation.

Recent works explore the task of language-guided dexterous grasp generation\cite{li2024multi, li2024semgrasp, zhang2024nl2contact, jian2025g-dex}, aiming to enable dexterous hands to perform actions based on language instructions, going beyond previous works that focus on stable grasping \cite{wang2023dexgraspnet, liu2020deep, shao2020unigrasp}. The typical language-guided approaches employ language as the condition of generative models to predict hand parameters, achieving impressive performance on objects within known categories \cite{wei2025grasp, chang2024text2grasp}. However, in the open real world, there are many categories that may not appear during training, and the cost of collecting data for dexterous hands is quite expensive. Therefore, the open-set generalization on unseen category samples is crucial for robot grasp. While previous works on parallel grippers explore open-set tasks \cite{rashid2023lerf-tog, li2024shapegrasp, DBLP:journals/corr/abs-2505-15660}, there is limited research on open-set generalization for dexterous hands, as their significantly higher degrees of freedom present complex challenges.

In this work, we explore a novel and challenging task: open-set language-guided dexterous grasping, as shown in Figure \ref{first_figure}, where models are evaluated on objects and language instructions from both seen and unseen categories. This task poses a great challenge in ensuring that grasps are intention consistent with corresponding language instructions for unseen categories. We find that the main challenge lies in the huge gap between high-level natural language and low-level robot action spaces, which makes it difficult to generalize the ability of understanding intentions and grasping from the training domain to unseen categories.

To solve the above challenges, we propose the Affordance Dexterous Grasp (AffordDexGrasp) framework, with the insight of employing a new affordance as an intermediate representation to bridge the gap between high-level language and low-level grasp actions. Our affordance representation combines two characteristics: \textbf{generalizable} (generalizing to unseen categories based on language) and \textbf{instructive} (guiding dexterous grasp generation effectively). However, achieving these two characteristics is not trivial. For example, as shown in Figure \ref{fig: motivation}, fine-grained contact information can effectively guide grasp generation or optimization \cite{brahmbhatt2019contactdb, grady2021contactopt, jiang2021graspTTA, yang2021cpf, jian2023affordpose}, but it is difficult to generalize to unseen categories. In contrast, coarse information, such as object parts, can be obtained from a off-the-shelf model \cite{liu2023partslip, umam2024partdistill, peize2023vlpart}, but it is too coarse to guide dexterous hand actions with higher degrees of freedom. 

To achieve these characteristics, we propose the Generalizable-Instructive Affordance by defining a general dexterous affordance that aligns with category-independent information, such as intention, object parts, and direction. As shown in Figure \ref{fig: motivation}, the affordance represents the potential graspable regions of all grasps with the same semantics. In this way, the models do not need to learn complex dexterous contact patterns but instead focus on a general graspable area, which can be well aligned with category-agnostic semantic attributes and guide grasp generation effectively.

To generate affordance and employ it to guide grasp generation, our framework consists of two cascaded generative models. The Affordance Flow Matching generates affordance maps in a generalizable manner based on language. And the Grasp Flow Matching generates dexterous grasp poses under the effective guidance of affordance. Moreover, we introduce a pre-understanding stage and a post-optimization stage to further boost generalization. Specifically, we employ the Multimodal Large Language Model (MLLM) to pre-understand user intention to enhance generalization across diverse user commands. And we introduce an affordance-guided optimization to improve grasp quality while preserving consistency with user intentions.

For evaluating our framework, we build a open-set table-top language-guided dexterous grasp dataset, based on the language guided dexterous grasp dataset \cite{wei2025grasp, yang2022oakink}. We exclude specific categories from the training set to test the model's open-set generalization. Moreover, we provide high-quality rendered images to facilitate the usage by MLLM, and we extend the dataset to scene-level data to better simulate real-world environments. The comprehensive experiments are conducted on both the simulation open-set dataset and real-world environments. The results show that our framework can generate dexterous grasp with consistent intention and high quality in open set.

\section{Related work}
\subsection{Dexterous Grasp}
Dexterous grasping is critical in robotics, which equips robots with human-like grasping capabilities \cite{xing2025taccap, lin2025typetele}. Some methods \cite{liu2021synthesizing, zhang2024dexgraspnet2, wan2023unidexgrasp++, wang2024single} focus on grasp stability and quality, while recent studies explore task-oriented \cite{wei2024dextask, zhu2021dextask} or language-guided grasping \cite{li2024semgrasp, wei2025grasp} with specific semantic intention. For language-guided task, the data-driven methods achieve impressive performance in unseen object within seen categories by leveraging language conditioned generative model \cite{chang2024text2grasp, wei2025grasp}. However, we find that these models are hard to generalize for samples from unseen categories. And we find that the main challenge is the huge gap between language and grasp action. To solve this problems, we propose AffordDexGrasp framework with the insight that bridge the gap by generalizable-instructive affordance.

\begin{figure*}[t]
\centering
\includegraphics[width=\linewidth]{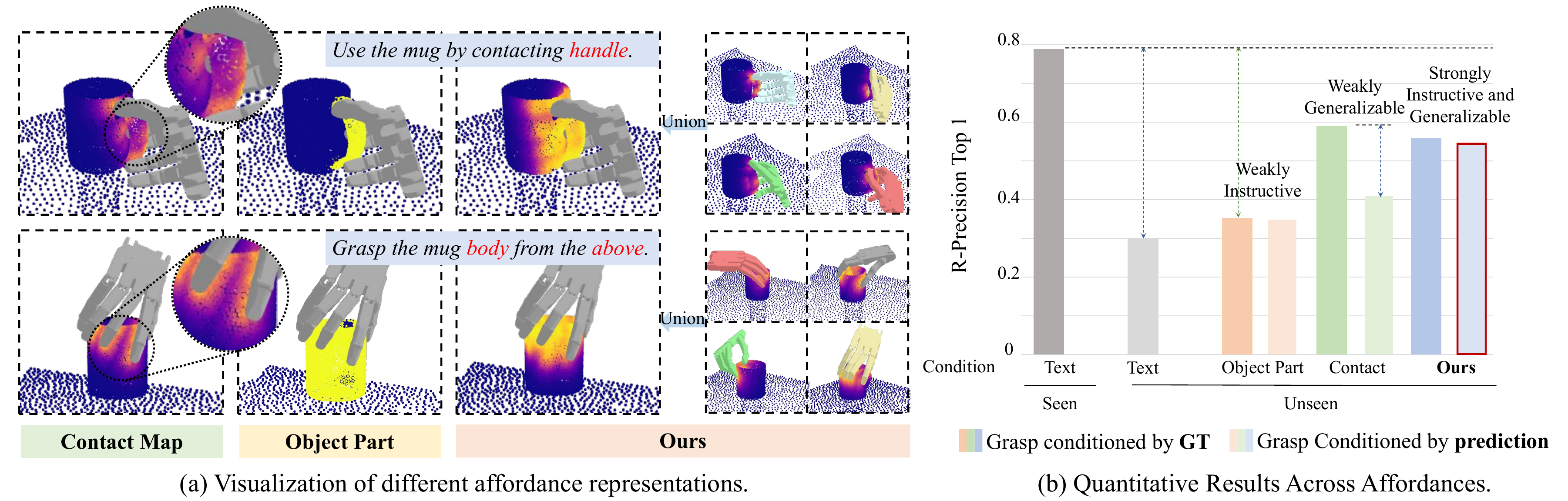}
\caption{
Different affordance representations. (a) While contact map are too elaborate to generalize and object part are too coarse to guide grasping, our affrodacne achieve a balance. (b) While object part has a lower upper bound and contact shows significant degradation in generalization, only our affordance effectively achieve the balance (Top-1 indicates grasp intention consistency).
}
\label{fig: motivation}
\end{figure*}

\subsection{Open-set Robot Grasp}
Exploring the performance of robotic models in open-set scenarios is crucial due to the diverse object categories in the real world and the high cost of data collection. For tasks that only focus on stable grasping, both parallel grippers \cite{fang2023anygrasp, wu2024economic, DBLP:journals/corr/abs-2502-16976} and dexterous hands \cite{wang2024unigrasptransformer, zhang2024dexgraspnet2} achieve impressive performance on open set. However, when it comes to considering task-oriented \cite{wang2025task} or language-guided grasping \cite{ma2024contrastive}, it becomes much more difficult. Some works in parallel grippers achieve this by using pre-trained visual grounding models \cite{li2022glip, peize2023vlpart, liu2023partslip}, implicit feature semantic fields \cite{kerr2023lerf, rashid2023lerf-tog, ma2024glover}, or multimodal large models \cite{huang2023voxposer, huang2024rekep}. Compared to parallel grippers, dexterous hands have a significantly higher number of degrees of freedom, making the methods of parallel grippers difficult to transfer to dexterous hand \cite{wei2025grasp}. In this paper, we explore the open-set language-guided dexterous grasping task and propose a novel framework to address it.

\subsection{Grasp Affordance}
Affordance is first introduced in \cite{gibson1979affordances}, referring to environmental action possibilities. There are currently two types of grasp affordance: fine-grained contact \cite{taheri2020grab, brahmbhatt2020contactpose} and coarse-grained object segmentation \cite{liu2023partslip}. The fine-grained information, such as contact areas, normals, or key points, is commonly used in hand-object interaction\cite{zhang2024nl2contact, zuo2024graspdiff, grady2021contactopt, brahmbhatt2019contactdb, zeng2025chainhoi}. However, this fine-grained information is difficult to generalize well to unseen categories. As a result, inaccurate contact information would lead to unreasonable grasping. On the other hand, We find through experiments that the coarse object segmentation \cite{liu2023partslip} is too coarse to guide dexterous grasp generation with relatively high degrees of freedom. To solve this problem, we propose Generalizable-Instructive Affordance, which can generalize through the local structure of objects using category-independent clues and effectively guide grasp generation.

\begin{figure*}[]
\centering
\includegraphics[width=\linewidth]{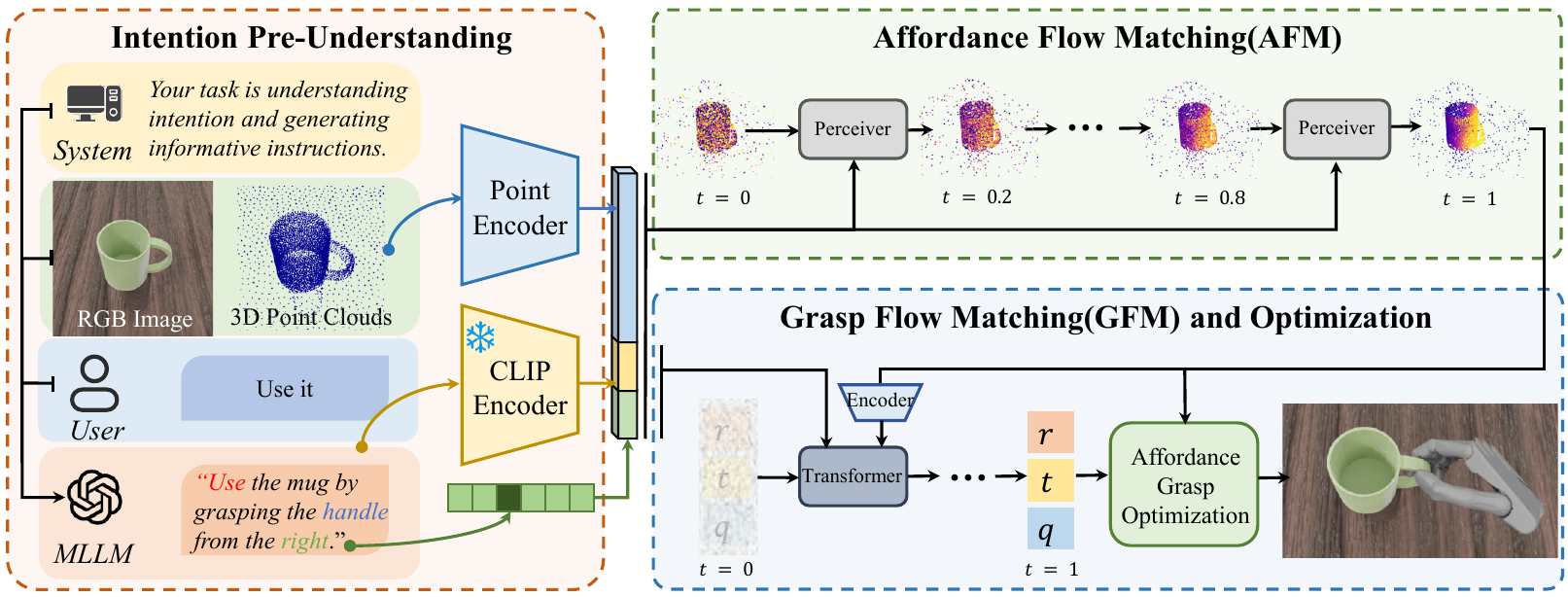}
\vspace{-0.5cm}
\caption{
The pipeline of Affordance Dexterous Grasp framework. The inference pipeline includes three stages: 1) intention pre-understanding assisted by MLLM; 2) affordance flow matching for generating affordance base on MLLM ouput; 3) Grasp Flow Matching and Optimization for outputing grasp poses based on the affordance and MLLM outputs. In the training time, AFM and GFM are independently trained one after another. Transformer and Perceiver are attention-based interaction module for velocity vector field prediction.}
\label{fig: methods}
\end{figure*}

\section{Affordance Dexterous Grasp Framework}
% \subsection{Framework Overall} 
Given the scene point cloud $\mathcal{O}$, RGB images $\mathcal{I}$ and language command $\mathcal{C}$ as input, our goal is to generate intention aligned and high-quality dexterous grasps poses $\mathcal{G}^{dex}$ = $(r, t, q)$, where $r$ denotes the rotation, $t$ denotes the translation, and $q$ represents the joint angles of the dexterous hand. 

In this section, we first introduce a novel generalizable-instructive affordance representation (Section \ref{Generalizable-Instructive Affordance}). Built on this affordance representation, we then propose the Affordance Dexterous Grasp framework (Section \ref{Affordance Dexterous Grasp framework}) including intention pre-understanding stage, Affordance and Grasp Flow Matching, and grasp post-optimization stage.

\subsection{Generalizable-Instructive Affordance} 
\label{Generalizable-Instructive Affordance}
We propose the generalizable-instructive affordance as an intermediate representation to bridge the gap between high-level language and low-level grasp actions. The affordances serves two primary objectives: 1) enable generalization by aligning object local structures with category-agnostic language semantics; and 2) provide instructive guidance for grasp generation through object affordance cues. However, there exists a trade-off in the generalization and instruction, and it is not trival to achieve both objectives.

\noindent\textbf{Generalization vs. Instruction.} While more fine-grained contact information can better guide grasp generation, it becomes harder to generalize to unseen categories. Fine-grained affordances typically compute the distance between object points and predefined elements, such as hand surface points or hand key points \cite{liu2023contactgen, li2022contact2grasp}. However, we observe that these fine-gained representations are difficult to generalize to unseen category samples as shown in Figure \ref{fig: motivation}. As the dexterous hand structure is complex and contact modes are varied, the contact maps are also elaborate and varies in space, which make it difficult to generalize. 

On the other hand, the information agnostic to dexterous hands, like object part segmentation, can be obtained in a generalizable manner by vision models \cite{liu2023partslip, zhou2023partslip++} or semantic feature fields \cite{rashid2023lerf-tog}. However, the object part is too coarse for dexterous hand, as it may struggle to define the clear grasping area. For example, grasping the body of a mug from above and from the side involves the same part, but with different intentions. Moreover, this coarse-grained learning may hinder the model from learning an accurate grasping method during training.

\noindent\textbf{Generalizable-Instructive Affordance.} 
To address this challenge, we propose a novel affordance representation that simultaneously enables both generalization and instruction. The key insight is to establish a correspondence between the general graspable affordance of objects and category-agnostic semantic attributes. Unlike contact maps that describe a particular hand contact area, our affordance represents all potential graspable regions that share the same semantic intention. This approach allows the affordance to generalize more effectively with category-agnostic semantic attributes from language, as it doesn't have to learn the complex distribution of hand contact. Meanwhile, our affordance serves as priors to enhance grasp generation by providing valuable graspable area cues.

To obtain the ground truth of generalizable-instructive affordance, we first organize those grasps of one object with the same intention, contact parts and grasp direction into a semantic group. Then we calculate the distances between each scene points $ \{o_i\}^M$ and hand surface points $\{h_i\}^N$, to obtain the contact map $\Omega_{contact} = \{d_i\}_1^M$ for each grasp in the group, where $d_i = \min\limits_{j = 1\dots N} \left\lVert o_i - h_j \right\rVert_2$. Then we unite contact areas of $k$ grasps in each group by calculating the minimum value of contact areas: ${d^{min}_i} =\{\min\limits_{k}(d_{i}^{k}) \}^M$. In order to make the affordance map more smooth, we apply Gaussian filtering to the union contact map: $a_i = \sum_{j \in \mathcal{N}(i)} \hat{w}_{ij} d_{j}^{min}$ and the weights are obtained by:
\begin{equation}
    \hat{w}_{ij} = \frac{\exp \left( -\| o_i - o_j \|^2 / (2\sigma^2) \right)}
    {\sum_{k \in \mathcal{N}(i)} \exp \left( -\| o_i - o_k \|^2 / (2\sigma^2) \right)},
\end{equation}
where $\mathcal{N}(i)$ is the neighborhood of $o_i$, $\sigma$ is set as the average nearest neighbor distance. Finally, we obtain the ground truth of generalizable-instructive affordance $\Omega = \{a_i\}^M$. 

\subsection{AffordDexGrasp Framework}
\label{Affordance Dexterous Grasp framework}

\subsubsection{Intention Pre-Understanding} 
\label{Intention Pre-Understanding}
The variability of user expression in the open world presents a challenge for models in generalizing and understanding user intentions. To address this, we employ the Multimodal Large Language Model (MLLM) to pre-understand user intentions, by imputing the user language and the rendered RGB image into GPT-4o \cite{hurst2024gpt4o} for intention understanding and key cues extraction. We empirically find that the following cues are both helpful for grasping and easily reasoned by the MLLM: the object category, user intention (e.g., use it), the contacting part of the object (e.g., the handle of a mug), and the grasp direction. The key cues can be extract from language when the commands are clear, otherwise can be reasoned from the image. 

To improve the understanding ability of MLLM, we employ chain of thought (CoT) and visual prompt, and the prompts can be found in supplementary material A.4. Additionally, the grasp direction is simplified as a discretized direction in six predefined coordinates (front, back, left, right, up, down). We first ask MLLM to obtain the direction in image coordinates, and transform it to world coordinates by the camera pose. Then the direction vector is derived from the index of the nearest coordinate axis to form discretized direction. Finally, MLLM organizes this information into a concise sentence (e.g.,``use the mug from the left by contacting the handle"). This compact and information-dense sentence structure allows subsequent models to better understand users' intentions.

\subsubsection{Affordance Flow Matching}
\label{Affordance Flow Matching}
We introduce Affordance Flow Matching (AFM) to learn the affordance distribution efficiently, achieving intention aligned and generalizable affordance generation. AFM is built on the conditional flow matching model \cite{liu2022flow1, albergo2022flow2, lipman2022flow3} aiming to learn the velocity vector field from the noised affordance map $\Omega_{0}$ to the target affordance map $\Omega_{1}$, and the objective loss of AFM could be:
\begin{equation}
\mathcal{L}_{AFM} = \mathbb{E}_{t, \Omega_{0} \sim p_0, \Omega_{1} \sim p_1} \left\| v_{\eta}(\Omega_t, t \vert c) - (\Omega_{1} - \Omega_{0}) \right\|^2,
\end{equation}
where $p_0$ is Gaussian distribution with zero mean and unit variance at time $t=0$, and $p_1$ is the target distribution as time $t=1$. $v_{\eta}(\Omega_t, t \vert c)$ is parameterized as a neural network with weights $\eta$ and $c$ is the input condition features. 

Specifically, the input scene object point cloud $\mathcal{O}$ is encoded by Pointnet++ \cite{qi2017pointnet++}, language is encoded by a pre-trained CLIP model \cite{radford2021clip}, and the direction vector and time embedding are encoded by the Multilayer Perceptrons (MLP). Due to the permutation invariance of point level affordance, the noisy affordance are concatenated with object features to aware the position and semantic information. Then all features are fed into a Perceiver IO Transformer, following \cite{jaegle2021perceiver2, wang2024move}. The noisy affordance features are served as input and output array, and the features of language and direction are served as latent array. The outputs of the decoder go through an MLP to obtain the prediction of the flow $v_{\eta}(\Omega_t, t \vert c) \in R^M$. Finally, in the inference time, we adopt mulitple steps $1/\Delta t$ following the equation:
\begin{equation}
\hat{\Omega}_{t+\Delta t} = \Omega_t + \Delta t \, v_{\eta}(\Omega_t, t \vert c_1), \quad \ t \in [0, 1).
\end{equation}

\subsubsection{Grasp Flow Matching}
\label{Grasp Flow Matching}
We introduce Grasp Flow Matching (GFM) to learn the grasp distribution, conditioned on the output of MLLM and AFM. GFM is also built upon the flow matching model, and the objective is to generate a parameterized dexterous hand pose $\mathcal{G}^{dex}$ = $(r, t, q)$, which represents the global rotation, translation, and joint angles. In the training time, grasp generation requires calculating explicit losses in 3D space, which involves obtaining the target dexterous poses, which go through forward kinematics to obtain hand configuration \cite{xu2024dgtr}. To obtain the target pose, we estimate it in one step during the training stage:
\begin{equation}
\hat{\mathcal{G}^{dex}_{1}} = \mathcal{G}^{dex}_t + (1-t)*v_{\theta}(\mathcal{G}^{dex}_t, t \vert c_2), \quad \ t \in [0, 1), 
\end{equation}
where the condition features $c_2$ are the concatenation of object affordance, language and direction features. And the object affordance features are obtained by feeding affordance with point clouds into PointNet++ \cite{qi2017pointnet++}.

The losses of grasp generation include 1) the grasp pose regression L2 loss; 2) the hand chamfer loss to minimize the discrepancies between the actual hand shapes; 3) the hand fingertip key point loss to minimize the distance between fingertip position. The loss function can be formulated as:
\begin{equation}
\mathcal{L}_{GFM} = \lambda_{pose}\mathcal{L}_{pose}+\lambda_{chamfer}\mathcal{L}_{chamfer}+\lambda_{tip}\mathcal{L}_{tip},
\end{equation}
where $\lambda_{pose}, \lambda_{chamfer}$ and $\lambda_{tip}$ are loss weights. 

The object penetration loss, which is used to penalize the penetration of hands into objects, is excluded from grasp generation training, as it leads to significant challenges in the training process \cite{wei2025grasp}. Employing the penetration loss can reduce object penetration, but negatively impacts both intention alignment and generation diversity. Therefore, to achieve semantic-aligned grasp generation, we exclude object penetration loss and introduce affordance-guided grasp optimization to reduce object penetration.

\begin{table*}[]
\centering
{
\small
\begin{tabular}{l|cccll|cll|lll}
\toprule
\multicolumn{1}{c|}{\multirow{3}{*}{}} & \multicolumn{5}{c|}{Intention} & \multicolumn{3}{c|}{Quality} & \multicolumn{3}{c}{Diversity} \\ \cline{2-12} 
\multicolumn{1}{c|}{} & \multirow{2}{*}{$FID \downarrow$} & \multirow{2}{*}{$CD \downarrow$} & \multicolumn{3}{c|}{$R-Precision \uparrow$} & \multirow{2}{*}{$Suc. \uparrow$} & \multicolumn{1}{c}{\multirow{2}{*}{$Q1 \uparrow$}} & \multicolumn{1}{c|}{\multirow{2}{*}{$Pen. \downarrow$}} & \multicolumn{1}{c}{\multirow{2}{*}{$\delta_{t}\uparrow$}} & \multicolumn{1}{c}{\multirow{2}{*}{$\delta_{r}\uparrow$}} & \multicolumn{1}{c}{\multirow{2}{*}{$\delta_{q}\uparrow$}} \\ \cline{4-6}
\multicolumn{1}{c|}{} &  &  & \multicolumn{1}{l}{$Top1$} & $Top2$ & \multicolumn{1}{l|}{$Top3$} &  & \multicolumn{1}{c}{} & \multicolumn{1}{c|}{} & \multicolumn{1}{c}{} & \multicolumn{1}{c}{} & \multicolumn{1}{c}{} \\ \bottomrule

\rowcolor{gray!20} % 设置灰色底纹 
\multicolumn{12}{l}{\textit{\textbf{Open Set A}}} \\
\midrule

\multicolumn{1}{l|}{ContactGen\cite{liu2023contactgen}} &0.428  &9.38  &0.164  &0.232  &0.289  &11.6\%  &1.71e-4  &1.09  &2.86   &18.4 &39.2  \\
\multicolumn{1}{l|}{Contact2Grasp\cite{li2022contact2grasp}} &0.426  &10.2  &0.257  &0.337  &0.398  & 16.5\% &0.0172  &0.668  &5.73   &4.84  &48.3 \\

\multicolumn{1}{l|}{GraspCVAE\cite{sohn2015cvae}} &0.378  &5.55  &0.309  &0.402  &0.469 &21.9\%  &0.0150  &0.709  &4.36   &4.60  &21.2  \\
\multicolumn{1}{l|}{SceneDiffuser\cite{huang2023diffusion}} &0.303  &5.02  &0.397  &0.473  &0.523  &29.1\% &0.0151   &0.487   &7.37  &\textbf{9.44}  &\textbf{68.9}    \\
\multicolumn{1}{l|}{DexGYS\cite{wei2025grasp}} &0.297  &4.63  &0.317  &0.401  &0.463  &44.2\% &0.0512  &0.362  &6.30   &6.79  &54.7  \\
\midrule
\multicolumn{1}{l|}{Ours} &\textbf{0.231}  &\textbf{3.81}  &\textbf{0.480}  &\textbf{0.588}  &\textbf{0.666}  &\textbf{45.1\%}   &\textbf{0.0531}  &\textbf{0.293}  &\textbf{7.48}   &6.98  &61.5  \\

\bottomrule
\rowcolor{gray!20} % 设置灰色底纹 
\multicolumn{12}{l}{\textit{\textbf{Open Set B}}} \\
\midrule
\multicolumn{1}{l|}{ContactGen\cite{liu2023contactgen}} &0.492 &8.95  &0.196  &0.257  &0.299  &6.04\%  &1.30e-4  &1.11  &3.26   &18.7  &45.5  \\
\multicolumn{1}{l|}{Contact2Grasp\cite{li2022contact2grasp}} &0.369 &11.9  &0.248  &0.318  &0.367 &16.2\%  &0.00801  & 0.798    &5.22   &4.63  &40.7   \\

\multicolumn{1}{l|}{GraspCVAE\cite{sohn2015cvae}} &0.365  &5.35  &0.297  &0.357  &0.403  &24.2\%  &0.00421  &0.738  &3.96  &5.47  &25.6  \\
\multicolumn{1}{l|}{SceneDiffuser\cite{huang2023diffusion}} &0.271  &6.50 &0.302  &0.355  &0.393  &30.7\% &0.00851  &0.683  &5.41  &6.13   &\textbf{72.9}  \\
\multicolumn{1}{l|}{DexGYS\cite{wei2025grasp}} &0.358  &3.40  &0.294  &0.358  &0.403 &35.2\%  &0.0220  &0.691   &5.40 & 6.11  &59.0   \\
\midrule
\multicolumn{1}{l|}{Ours} &\textbf{0.162} &\textbf{2.95}  &\textbf{0.532}  &\textbf{0.609}  &\textbf{0.655} &\textbf{38.9\%}  &\textbf{0.0352}   &\textbf{0.361} &\textbf{6.86}  &\textbf{6.84} &56.3  \\
\bottomrule
\end{tabular}
}
\caption{Results on open-set datasets compared with the SOTA methods.}
\label{table: sota}
\vspace{-1em}
\end{table*}

\normalsize
\subsubsection{Affordance-guided Grasp Optimization}
\label{Affordance-guided Grasp Optimization}
We introduce a non-parametric test-time adaptation (TTA), affordance-guided grasp optimization, to improve the grasp quality while maintaining intention consistency. Compared to learning-based TTA methods \cite{jiang2021graspTTA, wei2025grasp}, our non-parametric optimization does not suffer from performance degradation caused by out-of-domain open-set samples. 

We design optimization objectives based on that high-quality grasps should exhibit high grasp stability and intention alignment in grasping posture and contact areas. Specifically, our optimization objectives includes: 1) affordance contact loss for constraining hand contact candidate points in contact with the affordance area:
\begin{equation}
    % \begin{split}
    \mathcal{L}_{\text{aff-dist}} = \sum_{i} \mathbb{I} \left( (d(p^{c}_{i}) < \tau_1) \lor (d(p^{r}_{i}) < \tau_1) \right) \cdot d(p^{r}_{i}),
    % \text{where} \quad d(p) = \min_{o \in \{o_j \in \mathcal{O} \mid a_j > \tau_2\}} \| p - o_j \|_2,
    % \end{split}
\end{equation}
where $d(p) = \min_{o \in \{o_j \in \mathcal{O} \mid a_j > \tau_2\}} \| p - o_j \|_2$, and $o_j$ and $a_j$ represent the position and affordance values of object points, and $p^{c}_{i}$ and $p^{r}_{i}$ represent the hand contact candidates of the initial coarse hand and current refined hand. 2) affordance fingertip loss to keep the fingertip positions that are in contact with the affordance area unchanged. 

3) object penetration loss for punish hand-object penetration, 4) self-penetration loss to punish the penetration of hand it self. 5) joint limitation loss to punish out-of-limit joint angles. The optimization objectives can be formulated as: 
\begin{equation}
    \min_{\mathcal{G}^{dex}} (\lambda_1\mathcal{L}_{\text{aff-dist}} + \lambda_2\mathcal{L}_{\text{aff-finger}}+\lambda_3\mathcal{L}_{pen}+\lambda_4\mathcal{L}_{spen}+\lambda_5\mathcal{L}_{joint}).
\end{equation}

\section{Open-set Dexterous Grasp Dataset}
To support the evaluation of our framework, we build an open-set dexterous grasp dataset in table-top environment based on the language-guided dexterous grasp dataset \cite{wei2025grasp, yang2022oakink}. Our open-set dataset consists of 33 categories, 1536 objects, 1909 scenes, and 43,504 dexterous grasps for Shadow Hand \cite{shadowhand} and Leap Hand \cite{shaw2023leap}. 

\noindent\textbf{- Open Set Data Split.} To enable comprehensive evaluation, we construct Open sets A and B using two independent dataset splits, each with its own set of unseen categories. For each split, all categories are first labeled as either seen or unseen. Then, 80\% of the objects from the seen categories are used for training, while the remaining 20\% and all objects from the unseen categories are used for testing. 

\noindent\textbf{- Scene Construction.} To make our dataset more practical, we generate scene data by placing objects in a tabletop environment using Blenderproc \cite{denninger2023blenderproc2}. To prevent collisions between the hand and the table, objects are elevated using a shelf, and collision detection is performed to filter out invalid grasps. The physics engine is activated to ensure physically plausible object placements and grasping poses, enhancing dataset quality.

\noindent\textbf{- Scene Point-Cloud and Image Rendering.} The scene was captured using five RGB-D cameras: four positioned at elevated lateral angles and one directly above the object. The partial point clouds are merged into a global point cloud. To render realistic RGB images for MLLM processing, we use the texture generation model Paint3D \cite{zeng2024paint3d} to apply realistic textures to all objects in our dataset.

\begin{table*}[]
\centering
\setlength{\tabcolsep}{5pt}
{\small
\begin{tabular}{c|ccccc|ccccc}
\toprule

\multirow{2}{*}{} & \multicolumn{5}{c|}{\textit{\textbf{Open Set A}}} & \multicolumn{5}{c}{\textit{\textbf{Open Set B}}} \\ \cline{2-11} 
 & $FID \downarrow$ & $CD \downarrow$ & $Top1 \uparrow$ & $Q1 \uparrow$ & $Pen. \downarrow$ & $FID \downarrow$ & $CD \downarrow$ & $Top1 \uparrow$ & $Q1 \uparrow$ & $Pen. \downarrow$ \\ \midrule
\rowcolor{gray!20}
\multicolumn{11}{l}{\textit{\textbf{Necessity of Generalizable-Instructive Affordance}}} \\  
\multicolumn{1}{l|}{w/o affordance} &0.338  &5.09  &0.369  &0.0180  &0.569  &0.322  &4.44  &0.308  &\textbf{0.0194}  &\textbf{0.452}  \\ 
\multicolumn{1}{l|}{w object part (pred)} &0.313  &5.06  &0.320  &0.0138  &0.594  &0.316  &3.79  &0.344  &8.12e-3  &0.721  \\ 
\multicolumn{1}{l|}{w contact map (pred)} &0.320  &4.89  &0.380  &0.0131  &0.605  &0.286  &3.25  &0.408  &9.25e-3  &0.636  \\
\rowcolor{yellow!20}
\multicolumn{1}{l|}{w our affordance (pred)} &\textbf{0.242}  &\textbf{3.79} &\textbf{0.480}  &\textbf{0.0240}  &\textbf{0.501}  &\textbf{0.176}  &\textbf{2.76} &\textbf{0.538}  &0.0150  &0.612  \\
\midrule
\multicolumn{1}{l|}{w object part (GT)} & 0.306 &5.05  &0.348  &0.0198  &0.523  &0.271  &3.79  &0.353  &6.68e-3  &0.526  \\ 
\multicolumn{1}{l|}{w contact map (GT)} &0.113  &2.01  &0.598  &0.0285  &0.494  &0.138  &2.68  &0.594  &0.0143  & 0.643  \\ 
\multicolumn{1}{l|}{w our affordance (GT)} &0.169  &3.52  &0.509  &0.0181  &0.545  &0.164  &2.61  &0.550  &0.0120  &0.584  \\
\midrule

\rowcolor{gray!20}
\multicolumn{11}{l}{\textit{\textbf{Effectiveness of Pre-Understanding Stage}}} \\  
\multicolumn{1}{l|}{w/o key cues extraction} &0.258  &3.99  &0.463  &0.0222  &0.510  &0.186  &3.42  &0.515  &8.33e-3  &0.702  \\  
\multicolumn{1}{l|}{w/o direction} &0.254  &4.07  &0.432  &0.0162  &0.531  &0.185  &2.85  &0.529  &8.80e-3  &0.679  \\  
\rowcolor{yellow!20}
\multicolumn{1}{l|}{w MLLM} &\textbf{0.242}  &\textbf{3.79} &\textbf{0.480}  &\textbf{0.0240}  &\textbf{0.501}  &\textbf{0.176}  &\textbf{2.76} &\textbf{0.538}  &\textbf{0.0150}  &\textbf{0.612}  \\ 
\midrule

\rowcolor{gray!20}
\multicolumn{11}{l}{\textit{\textbf{Effectiveness of Affordance-based Optimization}}} \\ 
\multicolumn{1}{l|}{w/o our optimization} &0.242  &\textbf{3.79} &\textbf{0.480}  &0.0240  &0.501  &0.176  &\textbf{2.76} &\textbf{0.538}  &0.0150  &0.612   \\ 
\multicolumn{1}{l|}{w penetration loss} &0.336  &7.07  &0.318 &0.0505  &0.299  &0.241  &5.76  &0.445  &0.0147  &0.719   \\ 
\multicolumn{1}{l|}{w ContactNet \cite{jiang2021graspTTA}} &0.349  &12.3  &0.279  &0.0434  &\textbf{0.124}  &0.355  &11.1  & 0.240 &0.0206  &\textbf{0.129}  \\ 
\multicolumn{1}{l|}{w RefineNet \cite{wei2025grasp}} &0.249  &3.96  &0.436  &0.0487  &0.399  & 0.203 &2.92  &0.493  &0.0149  &0.657  \\ 
\rowcolor{yellow!20}
\multicolumn{1}{l|}{w our optimization } &\textbf{0.231}  &3.81  &0.477  &\textbf{0.0531}  &0.293  &\textbf{0.162} &2.95  &0.532  &\textbf{0.0350}  &0.361  \\
\bottomrule
\end{tabular}
}
\caption{Ablation study for our framework. The results of first two experiment groups are obtained from model outputs without optimization for an intuitive and reasonable comparison. The results in each group should be compared with \colorbox{yellow!20}{\textit{light yellow line} (our default setting)}.}
\vspace{-0.2cm}
\label{table:AB}
\end{table*}

\section{Experiment}

\subsection{Evaluation Metrics} 
We employ three types of metric to evaluate the ability of intention consistency, grasp quality and grasp diversity. 

\noindent\textbf{- Evaluating Intention Consistency.} We employ several metrics, including FID, R-Precision and Chamfer Distance. 1) \textbf{FID} (Frechet Inception Distance) \cite{heusel2017fid} measures the distribution similarity between generated grasps and ground truth. To extract grasp and instruction features, we train an object-grasp point cloud encoder and a language encoder by contrastive learning \cite{guo2022fid_motion}. 2) \textbf{R-Precision} evaluates the semantic alignment between language instructions and generated grasps \cite{guo2022fid_motion}. Specifically, for each generated grasp, we construct a pool of 32 samples, including its ground-truth instruction and randomly selected samples. We then compute and rank the cosine distances between the point cloud features and the language features of all samples in the pool. The average accuracy is computed at the top-1, top-2, and top-3 positions, that a retrieval is considered successful if the ground truth entry appears among the top-k candidates. 3) \textbf{Chamfer Distance (CD)} quantifies the discrepancy between the predicted hand point cloud and the closest target with same intention \cite{wei2025grasp}.

\noindent\textbf{- Evaluating Grasp Quality.} Following \cite{wang2023dexgraspnet}, we use \textbf{success rate}, denoted as $Suc.$ , in Issac Gym \cite{liang2018isaac} and \textbf{Q1} \cite{ferrari1992q1} to assess grasp stability. \textbf{Maximal penetration depth} (\text{cm}), denoted as $Pen.$ is used to calculate the maximal penetration depth from the object point cloud to hand meshes. 

\noindent\textbf{- Evaluating Grasp Diversity.} We compute the \textbf{Standard deviation} of translation $\delta_t$, rotation $\delta_r$ and joint angle $\delta_q$ of samples within the same scene observation. 

\subsection{Implementation Details} 
For training our framework, the number of epochs is set to 50 for AFM and 200 for GFM. 
For AFM, the loss weight of L2 loss is set to 1. For GFM, we set $\lambda_{\text{pose}} = 10$, $\lambda_{\text{chamfer}} = 1$, and $\lambda_{\text{tip}} = 2$. The batch size is set to 16 for AFM and 64 for GFM. The initial learning rate is $2.0 \times 10^{-4}$, decaying to $2.0 \times 10^{-5}$ using a cosine learning rate scheduler~\cite{loshchilov2016sgdr}. The Adam optimizer is used with a weight decay rate of $5.0 \times 10^{-6}$. In the inference, the time step is set to 10 for AFM and 20 for GFM. For the optimization, $\lambda_1 = 100$, $\lambda_2 = 10$, $\lambda_3 = 100$, $\lambda_4 = 10$, $\lambda_5 = 100$. The number of optimization iterations is set to 200. All experiments are implemented using PyTorch on a single RTX 4090 GPU.

\subsection{Comparison with SOTA Methods}
We compare our methods with the SOTA methods, as shown in Table \ref{table: sota}. The generic grasp methods are reproduced by concatenating the point cloud features and features of same language guidance with ours. The same encoders are employed for fair comparison, and the penetration loss is excluded to avoid learning difficulties according to \cite{wei2025grasp}. The results show that our framework significantly outperforms all previous methods in terms of intention consistency and grasp quality. Our framework also achieves a reasonably high level of diversity, as excessive diversity may lead to unnatural postures. Previous language-guided methods fail to generalize well to unseen categories due to the huge gap between language and grasping actions. Similarly, contact-based methods don’t perform well, as the contact maps are difficult to generalize. Additionally, our method outperforms other methods in the close set, as shown in Table \ref{table: close set}. Overall, the results indicate that our framework achieves strong performance in generating intention-aligned, high-quality, and diverse grasps.

\begin{figure*}[t]
\centering
\includegraphics[width=\linewidth]{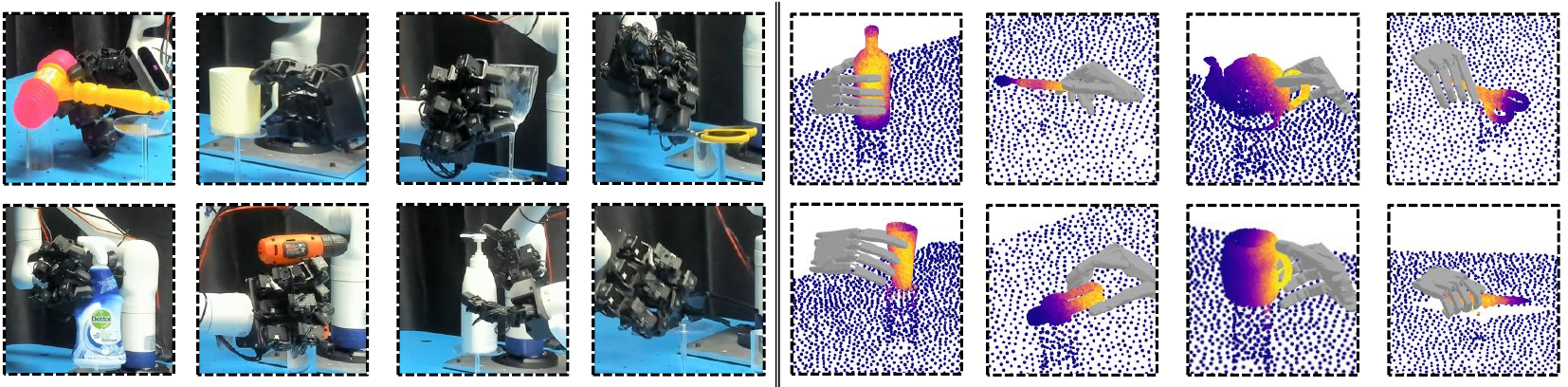}
\vspace{-2em}
\caption{
The visualization of generated affordance and dexterous grasp. The \textbf{left top} shows the zero-shot samples and the \textbf{left bottom} shows the one-shot samples in real world. The \textbf{right top} and \textbf{right bottom} show the zero-shot samples in simulation open set A and B.  
}
\label{fig: vis}
\end{figure*}

\begin{table}[ht]
\setlength{\tabcolsep}{3pt}
\centering
{\small
\begin{tabular}{c|ccccc}
\toprule

\multicolumn{1}{l|}{}  & $FID \downarrow$ & $CD \downarrow$ & $Top1 \uparrow$ & $Q1 \uparrow$ & $Pen. \downarrow$ \\
\midrule
% \multicolumn{1}{l|}{Contact2Grasp} &  &  &  &  &  \\ 
\multicolumn{1}{l|}{GraspCVAE} &0.208  &2.40  &0.395  &0.0100  &0.771  \\ 
\multicolumn{1}{l|}{DexGYS} &0.0804  &1.74  &0.590  &0.0551  &0.397  \\
\midrule
\multicolumn{1}{l|}{Ours} &\textbf{0.0286}  &\textbf{1.13}  &\textbf{0.779}  &\textbf{0.0562}  &\textbf{0.193}  \\ 
\bottomrule

\end{tabular}
}
\caption{Experiment of close set with SOTA methods.}
\vspace{-0.5cm}
\label{table: close set}
\end{table}

% and trade-off in 
\subsection{Necessity of Our Affordance}
The results shown in Table \ref{table:AB} validate the key insight of our framework: using the generalizable-instructive affordance to bridge the gap between high-level language and low-level grasp actions. The first row presents the results without affordance, which fails to generalize to unseen categories. The subsequent two rows show the results under the condition of object parts and contact maps. Prediction refers to maps generated by models, while GT refers to using the corresponding ground truth. As discussed in Section \ref{Generalizable-Instructive Affordance}, the results demonstrate that object parts are too coarse to guide the grasp effectively, even when using the ground truth. On the other hand, the contact map is too fine-grained to generalize to unseen categories. Only our affordance achieves a balance between generalization and instruction.

\begin{table}[ht]
\setlength{\tabcolsep}{2pt}
\centering
\begin{tabular}{c|cccccc}
\toprule
\multirow{2}{*}{} & \multicolumn{2}{c|}{Seen Cate} & \multicolumn{2}{c|}{Similar Cate} & \multicolumn{2}{c}{Novel Cate} \\ \cline{2-7} 
 & $Top1 \uparrow$ & \multicolumn{1}{c|}{$CD \downarrow$} & $Top1 \uparrow$ & \multicolumn{1}{c|}{$CD \downarrow$} & $Top1 \uparrow$ & $CD \downarrow$ \\ \midrule
\rowcolor{gray!10}
\multicolumn{7}{l}{\textit{\textbf{Zero-Shot}}} \\  
w/o afford &0.827  &2.61  &0.295  &5.71  &0.219  &7.82   \\
Ours &\textbf{0.880}  &\textbf{1.06}  &\textbf{0.432}  &\textbf{3.65}  &\textbf{0.246}  &\textbf{6.42}  \\ 
\midrule
\rowcolor{gray!10}
\multicolumn{7}{l}{\textit{\textbf{One-Shot}}} \\  
w/o afford &0.767  &3.17 &0.429  &3.67  &0.342  &6.64  \\
Ours &\textbf{0.870}  &\textbf{0.99} &\textbf{0.656}  &\textbf{1.73}  &\textbf{0.586}  &\textbf{3.03}   \\

\bottomrule
\end{tabular}
\caption{Zero-shot and one-shot generalization of our framework.}
\label{table: one shot}
\vspace{-0.2cm}
\end{table}

Furthermore, our affordance representation and framework demonstrate good generalization in the one-shot setting, as shown in Table \ref{table: one shot}. For the one-shot experiments, we introduce several novel categories to the test set that differ significantly from the training set and add one object from each unseen category to the training set. The results show that our framework achieves a significant performance improvement, not only on samples from different categories with similar structures but also for novel categories.

\subsection{Effectiveness of Each Component}
Further ablation studies are conducted to evaluate each component of our framework. For the pre-understanding stage assisted by MLLM, \textit{w/o key cues extraction} refers to directly inputting the user’s commands into the model, while \textit{w/o direction} means that the direction information is not utilized. Both designs enhance performance, and we believe that MLLM would be more beneficial in real-world applications due to its powerful generalization. For affordance-based optimization, our optimization improves grasp quality and maintains intention consistency, outperforming the learning-based methods ContactNet \cite{jiang2021graspTTA} and RefineNet \cite{wei2025grasp}, as well as the model which is trained with penetration loss.

\subsection{Real World Experiments}
The real-word experiments are conducted to verify the simulation-to-reality ability of our framework. We employ a Leap Hand, a Kinova Gen3 6DOF arms and an original wrist RGB-D camera of Kinova arm. And we collect several common objects in daily life, including unseen objects and unseen categories. In experiment, we synthesize the scene point cloud by taking several partial depth maps around the object. Then the scene point cloud, a RGB image and the user language command are fed into our framework to obtain the dexterous grasp pose. During execution, we first move the the arm to a pre-grasp position, then synchronously move the joints of the robotic arm and the dexterous hand to reach the target pose. In evaluation, a grasp is considered as success if the hand can lift it up and the action is consistent with the given command. The results shown in Table \ref{table:RealWord} shows that our framework achieve higher success rate than previous method.

\begin{table}[]
\centering
\begin{tabular}{ccccccccc}
\toprule
\multicolumn{1}{l|}{} 
& \raisebox{-0.35\height}{\includegraphics[width=0.05\linewidth]{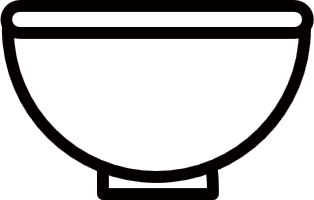}}
& \raisebox{-0.35\height}{\includegraphics[width=0.05\linewidth]{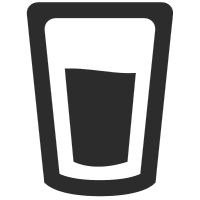}} 
& \raisebox{-0.34\height}{\includegraphics[width=0.05\linewidth]{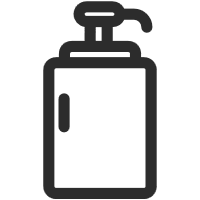}} 
& \raisebox{-0.35\height}{\includegraphics[width=0.05\linewidth]{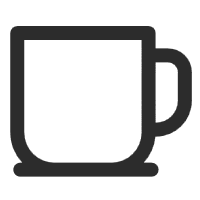}}
& \raisebox{-0.35\height}{\includegraphics[width=0.05\linewidth]{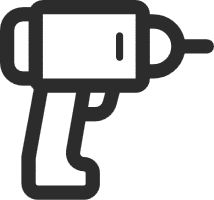}}
& \raisebox{-0.35\height}{\includegraphics[width=0.05\linewidth]{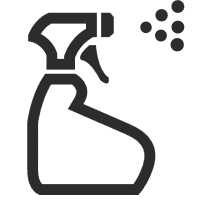}}
& \raisebox{-0.35\height}{\includegraphics[width=0.05\linewidth]{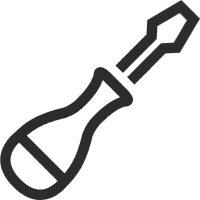}}
& \raisebox{-0.35\height}{\includegraphics[width=0.05\linewidth]{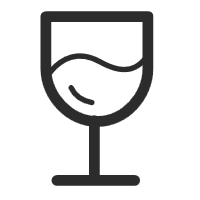}}
\\ \midrule
\multicolumn{1}{l|}{DexGYS} & 2 & \textbf{7} & 3 & 6 & 4 & 7 & 0 & 10 \\
\multicolumn{1}{l|}{Ours} & \textbf{10} & 5 & \textbf{9} & \textbf{10} & \textbf{6} & \textbf{10} & \textbf{6} & \textbf{10} \\ \bottomrule
\end{tabular}
\caption{Real world experiments: Successes in 10 attempts.}
\label{table:RealWord}
\vspace{-1em}
\end{table}

\section{Conclusions}
{
\linespread{1.001}\selectfont
We believe that achieving generalizable dexterous grasps in open set is crucial within the deep learning and robotics communities. In this paper, we explore a novel task of open-set language-guided dexterous grasp. This task is challenging due to the huge gap between language and grasping actions, which hinders the model’s generalization ability. We propose the AffordDexGrasp framework with the insight that using a new affordance representation, generalizable-instructive affordance, to solve this challenge. This affordance enables generalization to unseen categories by utilizing the object's local structure and category-agnostic semantic attributes, thus facilitating effective dexterous grasp generation. Based on this affordance, our framework introduces two flow matching based models for affordance generation and affordance-guided grasp generation. Moreover, we introduce a pre-understand stage and a pose-optimization stage to further boost generalization. We conduct extensive open-set experiments in both simulation and the real world, and the results show that our framework outperforms all SOTA methods.

}

\section{Acknowledgements}
This work was supported partially by NSFC (92470202, U21A20471), Guangdong NSF Project (No. 2023B151504 0025), Guangdong Key Research and Development Program (No. 2024B0101040004).

{
    \small
    \bibliographystyle{ieeenat_fullname}
    \bibliography{main}
}

\appendix
\newpage

\twocolumn[{
\begin{center}
    \LARGE\bfseries Supplemental Material
\end{center}
\vspace{1em}
}]

\begin{figure*}[t]
\centering
\includegraphics[width=\linewidth]{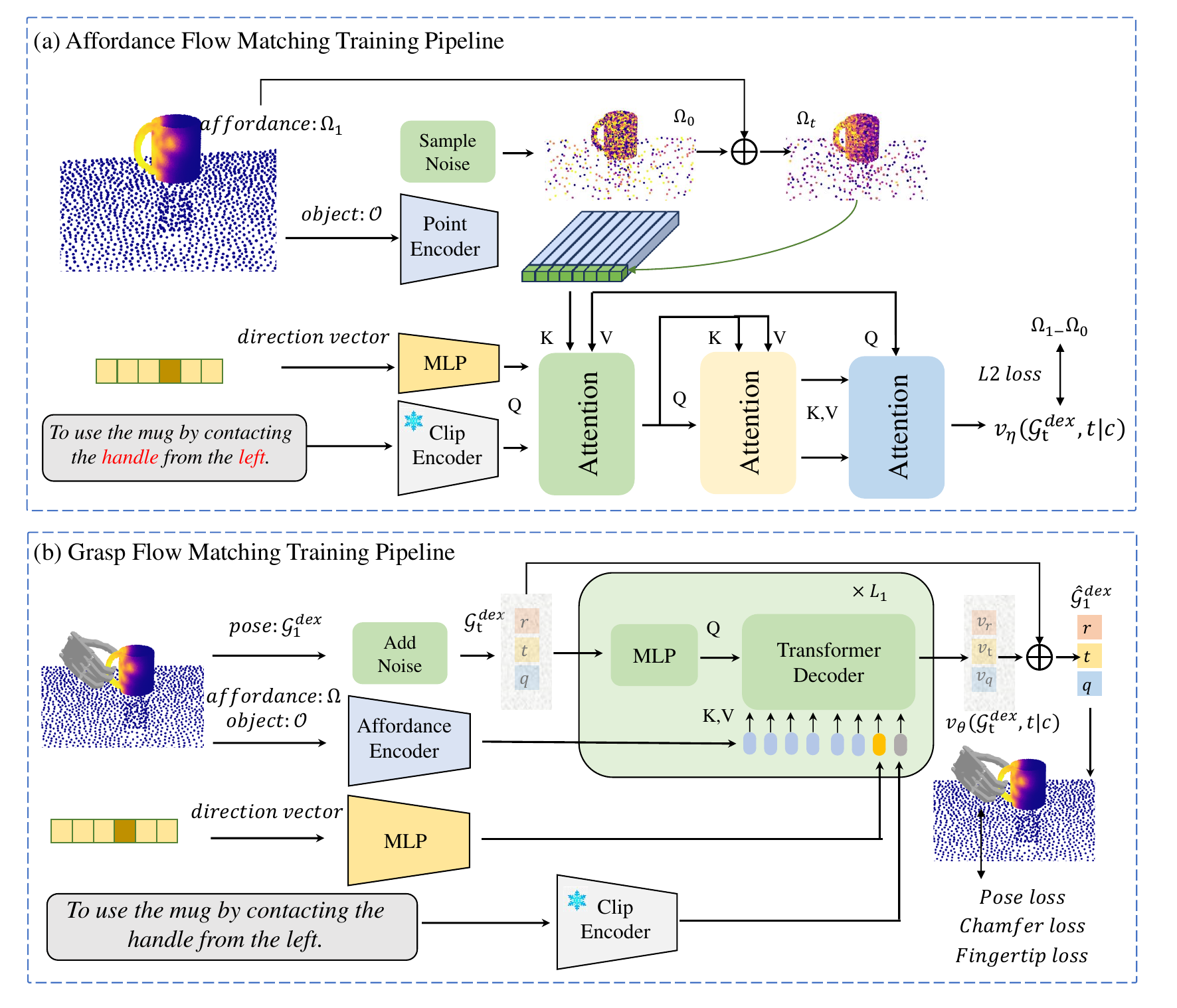}
\caption{The training pipeline of Affordance Flow Matching and Grasp Flow Matching.}
\label{fig: methods2}
\end{figure*}

\begin{figure*}[t]
\centering
\includegraphics[width=\linewidth]{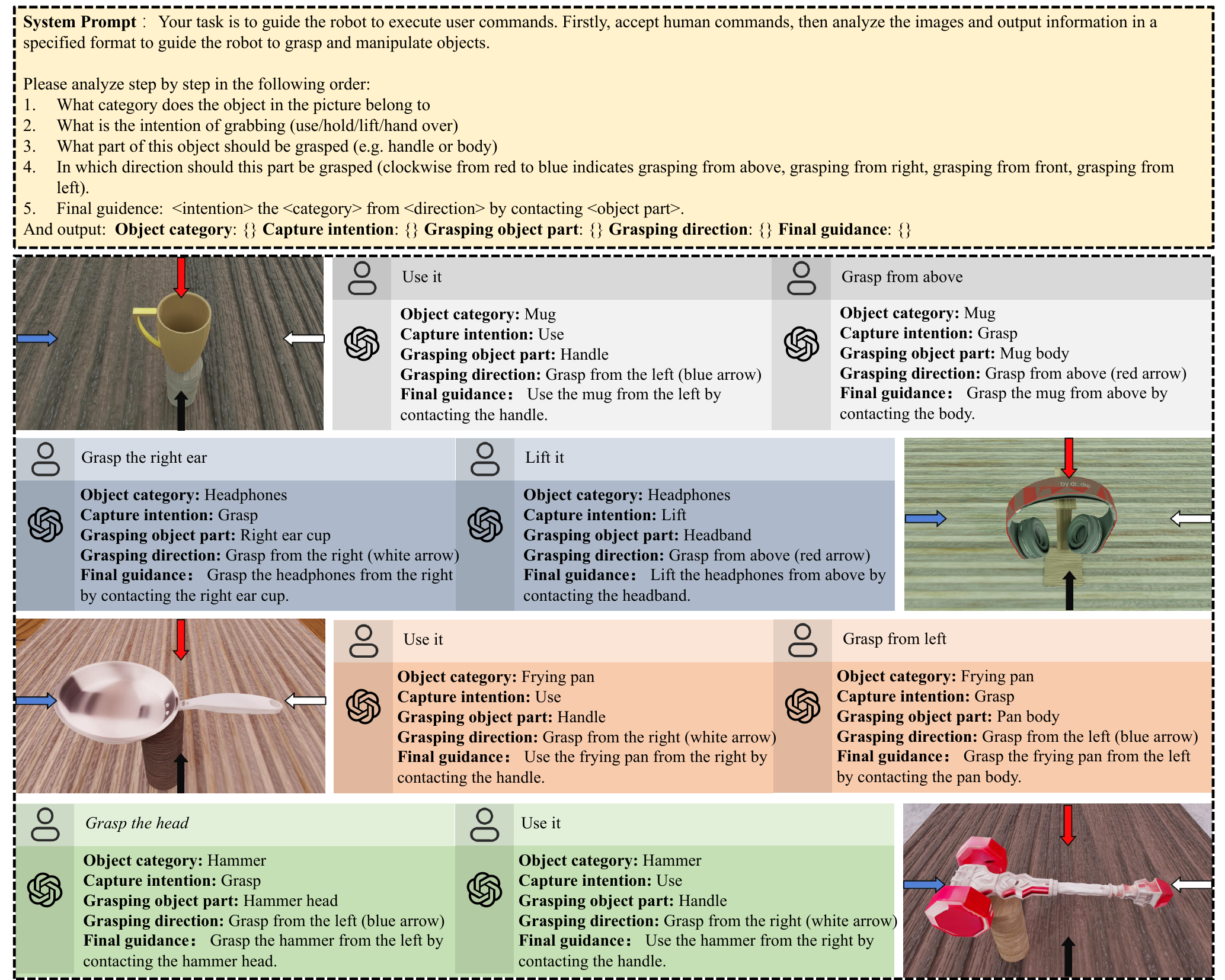}
\caption{The prompt used in per-understanding stage.}
\label{fig: mllm}
\end{figure*}

\begin{figure*}[t]
\centering
\includegraphics[width=\linewidth]{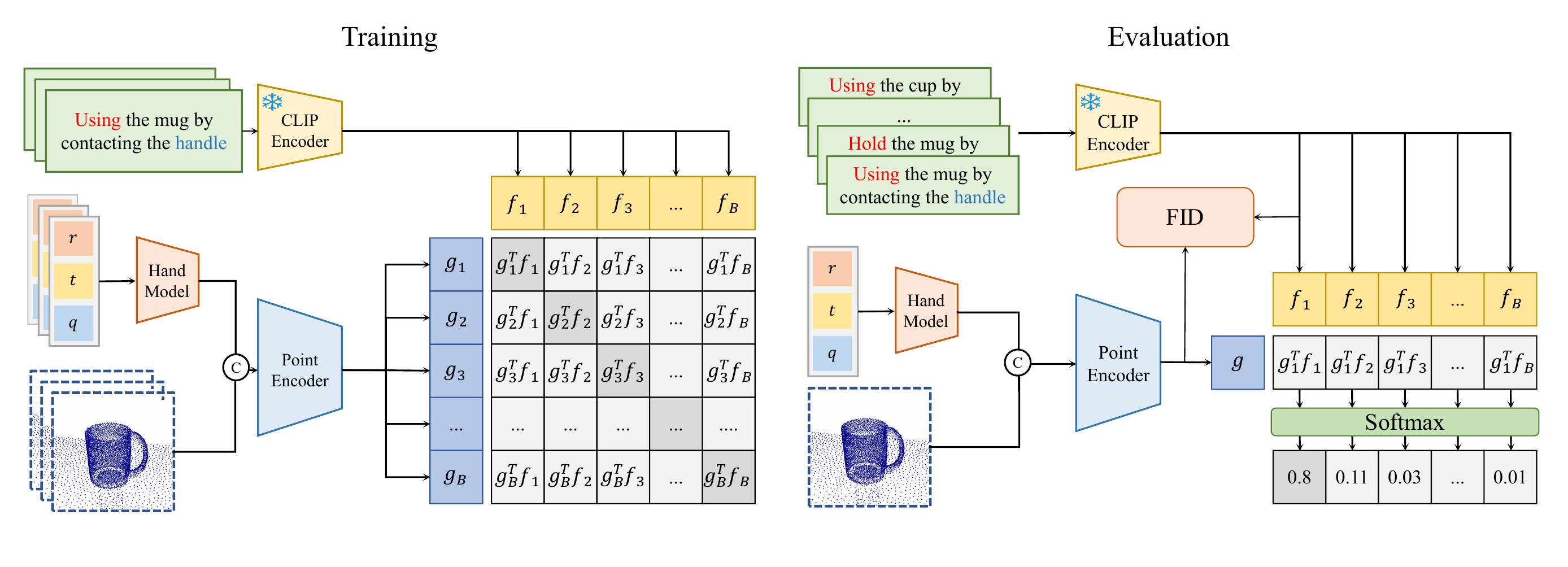}
\caption{The training pipeline of evaluation encoder (left). The evaluation of FID and R-Precision.}
\label{fig: eval}
\end{figure*}

\section{Framwork Details}

\subsection{Affordance Flow Matching}
The training pipeline of Affordance Flow Matching (AFM) is shown in Figure \ref{fig: methods2} (a). A noise $\Omega_{0}$ is first sampled from a Gaussian distribution with zero mean and one unit. A timestep $t$ is sampled from 0 to unit variance. Then, the training input $\Omega_{t}$ is obtained by $\Omega_{t} = (1-t) * \Omega_{0} + t * \Omega_{1}$, where $\Omega_{1}$ is the ground truth of the affordance. The scene point clouds $\mathcal{O}$ are fed into PointNet++ \cite{qi2017pointnet++} with several upsampling and downsampling layers to obtain features $\mathcal{F}_{obj} \in \mathbb{R}^{N_{scene} \times C_{scene}}$. The object features are then concatenated with $\Omega_{t}$ and point cloud positions $\mathcal{O}$, and processed by an MLP layer to obtain affordance features $\mathcal{F}_{aff} \in \mathbb{R}^{N_{scene} \times C_{aff}}$. The direction vector, a one-hot vector of 6 dimensions, is fed into an MLP to obtain the direction features $\mathcal{F}_{dir} \in \mathbb{R}^{1 \times C_{dir}}$. The language text is encoded by the CLIP model \cite{radford2021clip} and outputs language features $\mathcal{F}_{lan} \in \mathbb{R}^{1 \times C_{lan}}$. These features are then fed into Perceiver IO \cite{jaegle2021perceiver1, jaegle2021perceiver2} for effective feature interaction. Specifically, the direction and language features serve as the query, while the affordance features serve as the key and value for cross-attention. Then, the output of the first attention mechanism performs self-attention. Then, the scene features serve as the query and the self-attention result serve as key and value in the final cross-attention to obtain the final features. The features are then fed into an MLP to predict the flow $v_{\eta}(\Omega_t, t | c_1)$. The loss function is the L2 loss between predicted flow with the target:
\begin{equation}
\mathcal{L}_{AFM} =  \left\| v_{\eta}(\Omega_t, t \vert c) - (\Omega_{1} - \Omega_{0}) \right\|^2.
\end{equation}

\subsection{Grasp Flow Matching}
The training pipeline of Grasp Flow Matching (GFM) is shown in Figure \ref{fig: methods2}. The object and affordance are concatenated first then fed into PointNet++ with several down-sampling layer to obtain the affordance features. The direction and language are encoded same with AFM. And all features are concatenated in the token dimension to obatin the condition features. The noised grasp poses are obtained similar with AFM, which are encoded by MLP. Then the pose features serve as query, and the condition features serve as key and value in the transformer decoder. The output features are fed into an MLP to obtain the prediction pose flow. Then the grasp pose can be obtained by one step, during the training time:
\begin{equation}
\hat{\mathcal{G}^{dex}_{1}} = \mathcal{G}^{dex}_t + (1-t)*v_{\theta}(\mathcal{G}^{dex}_t, t \vert c_2), \quad \ t \in [0, 1), 
\end{equation}
The loss functions includes grasp pose regression loss, hand chamfer loss and hand fingertip key point loss, which are detailed in next section.

\subsection{Loss Function}
\label{sec: loss details}
This section provides a detailed exposition of the loss functions utilized during the grasp generation and optimization.

\noindent\textbf{- Parameter Regression Loss.}
We utilize the mean squared error (MSE) to quantify the deviation between the generated dexterous hand pose $\hat{\mathcal{G}^{dex}_1}$ and the ground truth $\mathcal{G}^{dex}_1$.
\begin{equation}
\mathcal{L}_{pose} = \|\mathcal{G}^{dex}_1 - \hat{\mathcal{G}^{dex}_1}\|_2^2.
\end{equation}

\noindent\textbf{- Hand Chamfer Loss.}
The predicted hand point clouds $\mathcal{H}(\hat{\mathcal{G}}^{dex}_1)$ and the ground truth $\mathcal{H}(\mathcal{G}^{dex}_1)$ are derived by sampling from the hand mesh. We then compute the chamfer distance to assess the discrepancies between the predicted and ground-truth hand shapes.

\begin{equation}
\label{chamferloss}
\begin{aligned}
    \mathcal{L}_{chamfer} 
    &= \sum_{x \in \mathcal{H}(\mathcal{G}^{dex})} \min_{y \in \mathcal{H}(\hat{\mathcal{G}}^{dex}_1)} \|x - y\|_{2}^{2} \\
    &\quad + \sum_{x \in \mathcal{H}(\hat{\mathcal{G}}^{dex})} \min_{y \in \mathcal{H}(\mathcal{G}^{dex}_1)} \|x - y\|_{2}^{2}.
\end{aligned}
\end{equation}

\noindent\textbf{- Fingertip Keypoint Loss.}
The fingertip keypoint loss $\mathcal{L}_{tip}$ measures the distance between the fingertip of generated grasp $q^g$ and the ground truth $q^{gt}$.
\begin{equation}
\mathcal{L}_{tip} = \|q^g - q^{gt}\|_2^2.
\end{equation} 

\noindent\textbf{- Affordance contact loss.}
The affordance contact loss is designed for constraining hand contact candidate points in contact with the affordance area:
\begin{equation}
    % \small
    % \begin{split}
    \mathcal{L}_{\text{aff-dist}} = \sum_{i} \mathbb{I} \left( (d(p^{c}_{i}) < \tau_1) \lor (d(p^{r}_{i}) < \tau_1) \right) \cdot d(p^{r}_{i}),
    % \text{where} \quad d(p) = \min_{o \in \{o_j \in \mathcal{O} \mid a_j > \tau_2\}} \| p - o_j \|_2,
    % \end{split}
\end{equation}
where $d(p) = \min_{o \in \{o_j \in \mathcal{O} \mid a_j > \tau_2\}} \| p - o_j \|_2$, and $o_j$ and $a_j$ represent the position and affordance values of object points, and $p^{c}_{i}$ and $p^{r}_{i}$ represent the hand contact candidates of the initial coarse hand and current refined hand. 

\noindent\textbf{- Affordance fingertip loss.}
The affordance fingertip loss is designed to keep the fingertip positions that are in contact with the affordance area unchanged. 
\begin{equation}
    \mathcal{L}_{\text{aff-finger}} = \sum_{i} \mathbb{I} \left( d({q}^{c}_{i} < \tau_3) \right) \cdot \| {q}^{r}_{i} - {q}^{c}_{i} \|_2
\end{equation}
where ${q}^{c}_{i}$ and ${q}^{r}_{i}$ represent the fingertip position of the initial coarse hand and current refined hand. 

\noindent\textbf{- Object Penetration Loss.}
The object penetration loss $\mathcal{L}_{pen}$ penalizes the depth of hand-object penetration, where $d_{i}^{sdf}$ denotes the signed distance from the object point to the hand mesh.
\begin{equation}
\mathcal{L}_{pen} = \sum_{i} \mathbb{I}(d_{i}^{sdf} > 0) \cdot d_{i}^{sdf}.
\end{equation}

\noindent\textbf{- Self-Penetration Loss.}
The self-penetration loss $\mathcal{L}_{spen}$ punishes the penetration among the different parts of the hand, where $p^{dex,sp}$ denotes predefined anchor spheres on the hand~\cite{xu2023unidexgrasp}.
\begin{equation}
\mathcal{L}_{spen} = \sum_{i,j} \mathbb{I}(i \neq j) \cdot \max(\delta - d(p_{i}^{dex,sp}, p_{j}^{dex,sp})).
\end{equation}

\noindent\textbf{- Joint Limitation Loss.}
Given the physical structure limitations of the robotic hand, each joint has designated upper and lower limits. The joint angle loss penalizes deviations from these limits.
\begin{equation}
\mathcal{L}_{joint} = \sum_{i} (\max(q_{i} - q_{i}^{max}) + \max(q_{i}^{min} - q_{i})).
\end{equation}

\noindent\subsection{Pre-understanding by MLLM}
We employ Chain-of-Thought (CoT) and visual prompt to enhance the reasoning ability of MLLM as shown in Figure \ref{fig: mllm}. Specifically, the MLLM is asked to reason the following task step by step: recognize the object category, understand the intention, reason the contact part based on intention and category, reason the direction based on the above. Futhermore, we add arrows of different colors in the four directions of the image and specified the direction represented by each arrow.

\section{Zero-shot Datasets Details}
\subsection{Dataset Splitting}
To support our Cross-Category Open-set settings, we divided the dataset into Open Set A and Open Set B by excluding specific samples from the training set. Specifically, the test set of Open Set A consists of 9,688 samples of 808 objects from 11 unseen categories: \{\textit{bowl}, \textit{cylinder}, \textit{wineglass}, \textit{headphones}, \textit{flashlight}, \textit{pen}, \textit{wrench}, \textit{toothbrush}, \textit{hammer}, \textit{teapot}, \textit{scissors}\}. The test set of Open Set B consists of 29,744 samples of 568 unseen objects from 10 unseen categories: \{\textit{bottle}, \textit{cup}, \textit{squeezable}, \textit{eyeglasses}, \textit{lightbulb}, \textit{fryingpan}, \textit{knife}, \textit{screwdriver}, \textit{mug}, \textit{pincer}\}. For one-shot experiments, we add an additional 3,688 samples of 152 objects from 12 categories: \{\textit{apple}, \textit{banana}, \textit{donut}, \textit{binoculars}, \textit{gamecontroller}, \textit{mouse}, \textit{phone}, \textit{stapler}, \textit{cameras}, \textit{lotion\_pump}, \textit{power drill}, \textit{trigger sprayer}\}.

\subsection{Texture generation}
As part of the texture generation, we utilized GPT-4o\cite{hurst2024gpt4o} to generate possible colors and materials for specific object categories. The prompt used for this process was as follows:

\textbf{Prompt}: ``Please generate a list of possible colors and materials for the following categories and provide the output in JSON format. The objects are: \{binoculars, mug, ...\}."

For objects lacking texture, we employ Paint3D\cite{zeng2024paint3d} to generate their visual appearance. The positive prompt is constructed using the object's category name along with a sampled color and material from the predefined possible lists, while the default negative prompt is utilized to guide the generation process.

\textbf{Prompt}: ``\texttt{<}\textit{category}\texttt{>}, \texttt{<}\textit{sampled color}\texttt{>}, \texttt{<}\textit{sampled material}\texttt{>}, high quality, clear"

\textbf{Negative prompt}: ``strong light, Bright light, intense light, dazzling light, brilliant light, radiant light, Shade, darkness, silhouette, dimness, obscurity, shadow"

\subsection{Scene Reconstruction}
\begin{figure*}[t]
\centering
\includegraphics[width=\linewidth]{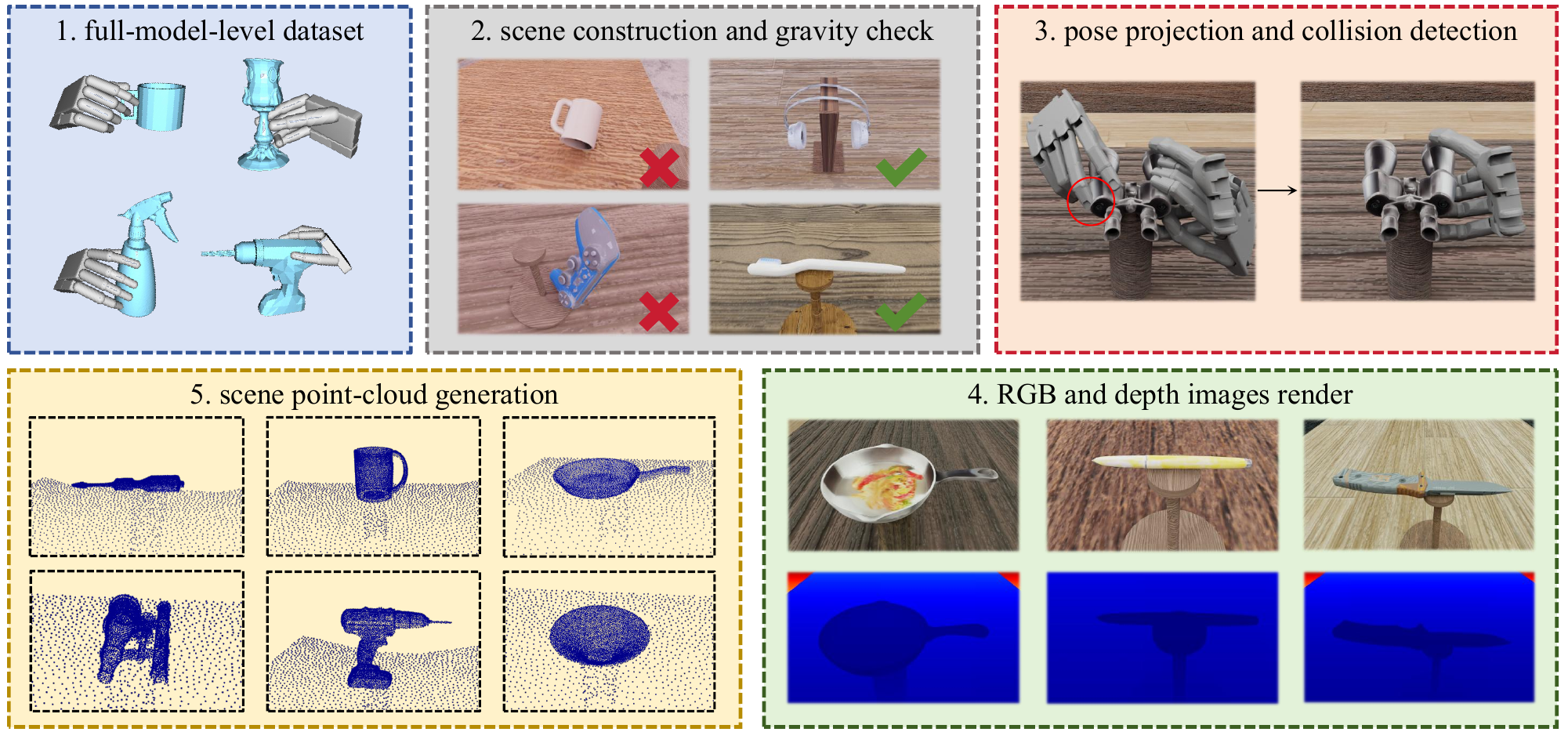}
\caption{The construction of our scene dataset.}
\label{fig: scenedataset}
\end{figure*}

The details of the construction of our scene dataset are illustrated in Figure~\ref{fig: scenedataset}. We utilize BlenderProc~\cite{denninger2023blenderproc2} to create a tabletop scene and project the grasping poses from the full-model dataset into the scene. After performing a gravity check on the objects and a collision test between the objects and the robotic hand, we record the grasp parameters in the scene coordinate system. At the same time, we capture RGB and depth images from the scene cameras. Finally, we construct the scene point cloud by concatenating and downsampling the partial point clouds obtained from the depth images.

In our tabletop scene, objects are lifted using three types of shelves. The first is a cylindrical base, which is used for the majority of categories. The second is a smaller shelf, consisting of two circular planes connected by a rod, designed for smaller categories such as pens and toothbrushes. The third is a headphone-specific shelf, used exclusively for headphones.
 
The poses of five RGB-D cameras are determined based on the center of the object's bounding box. One camera is positioned directly above the center of the bounding box. The other four cameras are placed at the four sides of the bounding box, with a side offset equal to half the length of the bounding box's longest edge and a height offset equal to one-fourth of the bounding box's longest edge.

\section{Experiments Details}
\subsection{Evaluation Details}
\textbf{- Training pipeline of evaluation encoder.}
The training pipeline of our Intention Evaluation Model is illustrated in Figure. \ref{fig: eval}. The model takes as input the hand parameters $\mathcal{G}^{dex}$, the object point cloud $\{o_i\}_{i=1}^{M}$, and the language guidance $\mathcal{L}$. Initially, the hand model generates the hand point cloud $\{h_i\}_{i=1}^{N}$, which is then concatenated with the object point cloud $\{o_i\}_{i=1}^{M}$ to form the combined hand-object point cloud $\{p_i\}_{i=1}^{P}$. 

A PointNet++ encoder and a CLIP encoder are subsequently employed to map $\{p_i\}_{i=1}^{P}$ and $\mathcal{L}$ into the feature space, producing the grasping feature $\mathbf{g}$ and the guidance feature $\mathbf{f}$, respectively. The Information Noise-Contrastive Estimation (InfoNCE) Loss\cite{oord2018representation} is then applied to maximize the cosine similarity between paired features while minimizing it between unpaired features. The InfoNCE Loss is defined as:
\begin{equation}
\mathcal{L}_{\text{InfoNCE}} = -\log \frac{\sum_{j=1}^B \mathbb{I}[\text{match}(i, j)] \exp(\text{sim}(\mathbf{g}_i, \mathbf{f}_j) / \tau)}
{\sum_{j=1}^B \exp(\text{sim}(\mathbf{g}_i, \mathbf{f}_j) / \tau)},
\end{equation}
where $\text{match}(i, j)$ indicates that the object category and action of the $i$-th grasp match those of the $j$-th grasp(including the case where $i=j$). Here, $\text{sim}(\mathbf{g}, \mathbf{f})$ denotes the cosine similarity between features $\mathbf{g}$ and $\mathbf{f}$, $\tau$ is a temperature parameter, and $B$ is the number of samples in the batch. The loss encourages high similarity for matched pairs $(\mathbf{g}_i, \mathbf{f}_j)$ and low similarity for unmatched pairs $(\mathbf{g}_i, \mathbf{f}_j)$ where $i \neq j$ and $\text{match}(i, j)$ does not hold.

To ensure the model's ability to distinguish between different actions within the same category, we design a custom sampler for the training data loader. The sampler randomly selects 8 different categories and randomly chooses $\frac{B}{8}$ grasps for each category. This process is repeated until the number of remaining categories in the dataset is fewer than 8.

\noindent\textbf{- R-precision}
When computing R-Precision during evaluation, we first randomly select 3 grasp guidance samples from the dataset that belong to the same category but involve different actions, along with 28 guidance samples from other categories. Next, we generate grasp features and a batch of guidance features using the same method as in the training stage. We then compute the cosine similarity between the grasp feature and each guidance feature. Finally, we determine the rank of the matched guidance sample based on these similarity scores.

\noindent \textbf{- Fréchet Inception Distance (FID).} This metric is widely used in generative tasks \cite{heusel2017fid, guo2022fid_motion} to assess the similarity between the generated and ground truth distributions. For our evaluation, we compute the FID using features extracted from the object and hand point clouds.

\noindent \textbf{- Chamfer Distance (CD).} Chamfer distance, denoted as $CD$, measures the discrepancy between the predicted hand point clouds and the nearest ground truth targets, providing a consistency measure from the perspective of hand positioning. The closest targets are the ground truth grasps that share the same intended grasp and are most similar to the generated grasp. This methodology follows the approach in \cite{wei2025grasp}.

\begin{figure*}[t]
\centering
\includegraphics[width=0.6\linewidth]{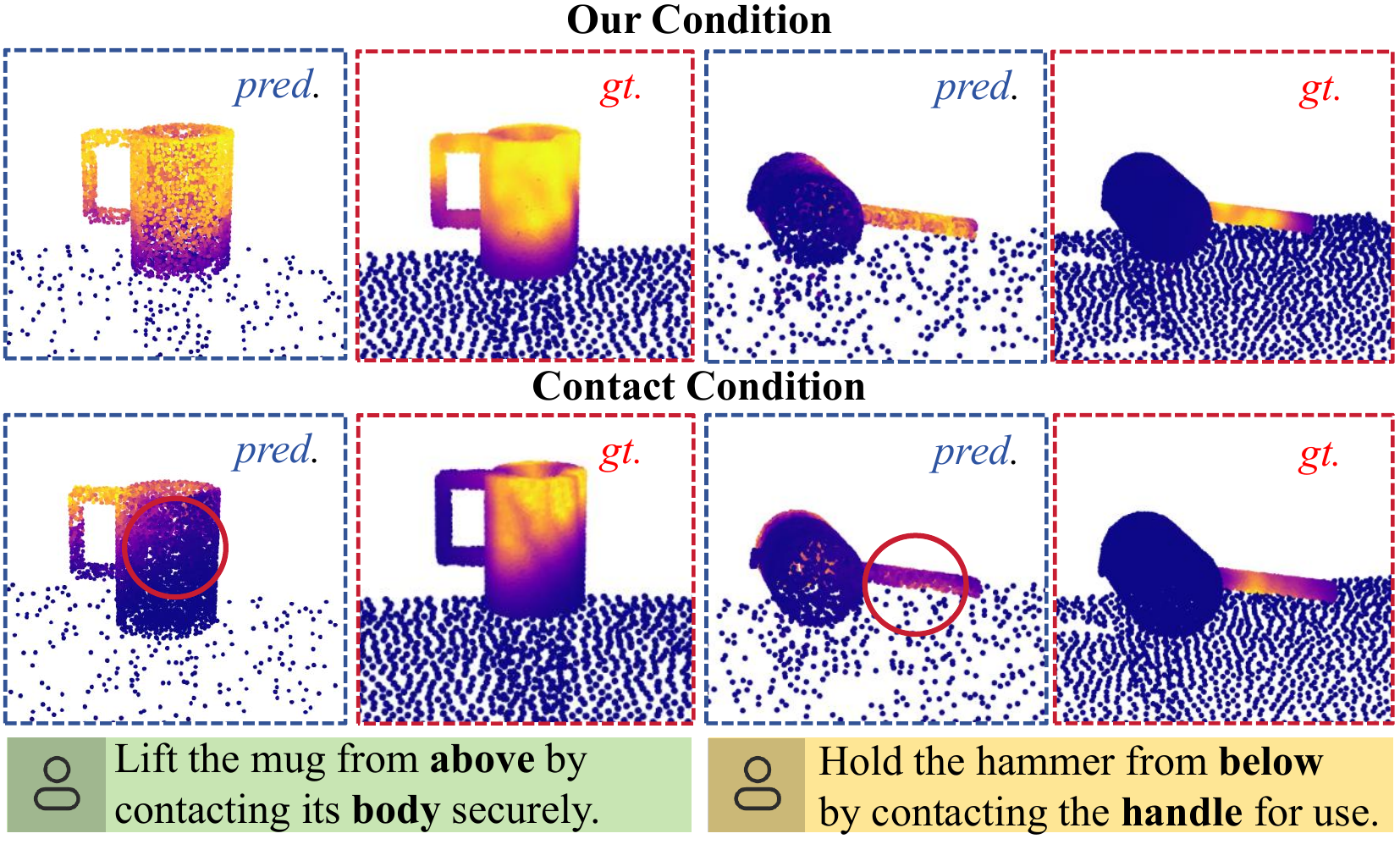}
\caption{The visualization of our affordance and contact map for prediction and the ground truth. }
\label{fig: affcompare}
\end{figure*}

\begin{figure*}[t]
\centering
\includegraphics[width=0.7\linewidth]{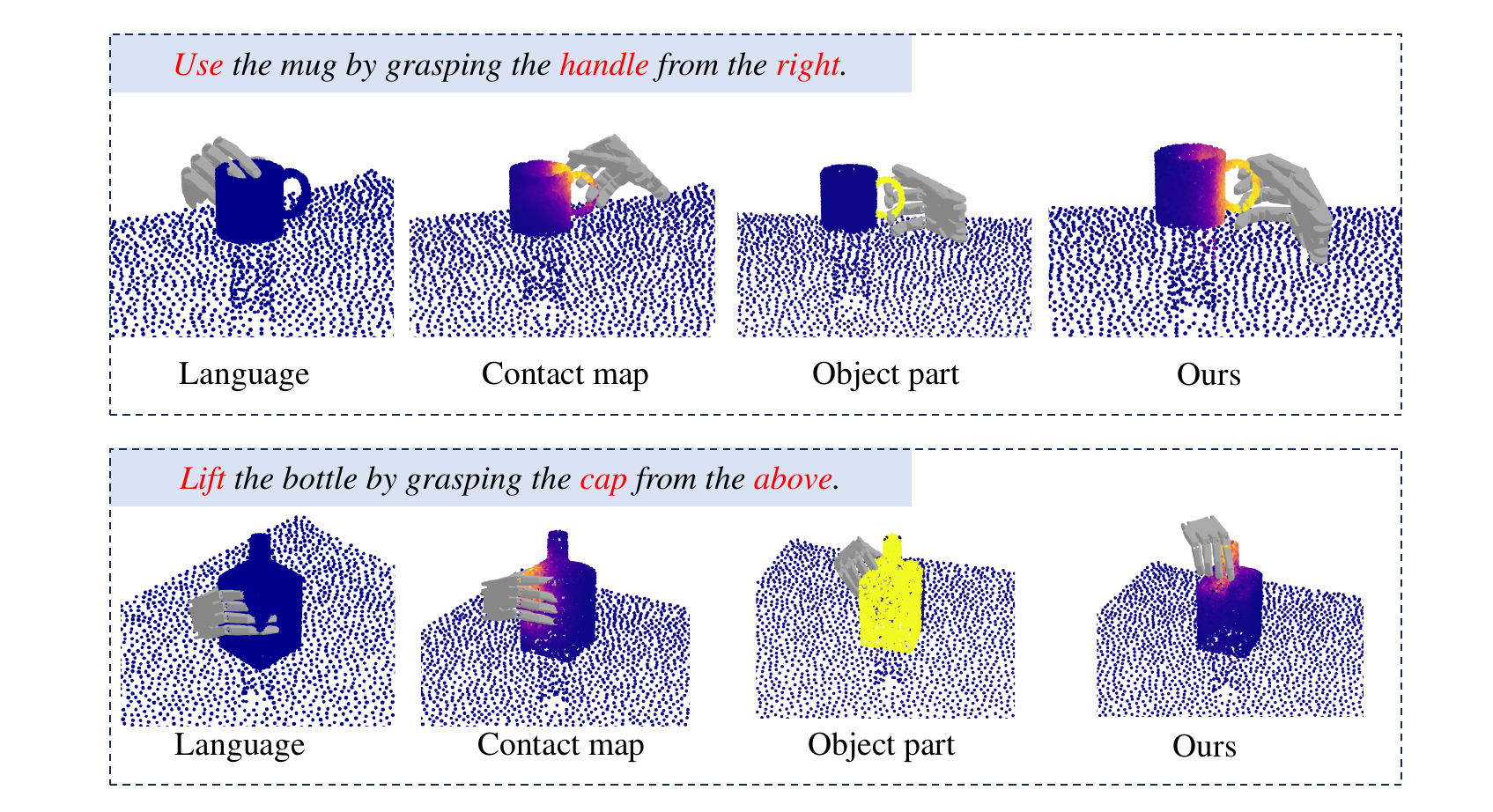}
\caption{The visualization of generated grasps under different condition with \textit{only language}, \textit{contact map}, \textit{object part} and \textit{ours}}.
\label{fig: graspcompare}
\end{figure*}

\noindent \textbf{- Grasp Success Rate.} The success rate of grasps is evaluated within the Isaac Gym simulation environment. To mimic the forces exerted by a dexterous hand in real-world scenarios, each finger is contracted towards the object. A grasp is considered successful if it can resist at least one of the six gravitational directions, indicating it can sustain a stable hold on the object.

\noindent \textbf{- Mean $Q_{1}$ Metric.} The $Q_{1}$ metric quantifies the smallest wrench capable of destabilizing a grasp. We follow the guidelines in \cite{wang2023dexgraspnet}, where the contact threshold is set at 1 cm and the penetration threshold is set at 5 mm. Any grasp with a maximal penetration depth exceeding 5 mm is considered invalid, and its $Q_{1}$ value is set to 0 before averaging the results.

\noindent \textbf{- Maximal Penetration Depth.} This metric measures the greatest penetration from the object’s point cloud to the hand mesh during a grasp.

\noindent \textbf{- Diversity.} To assess the diversity of generated grasps, we calculate the standard deviations of translation ($\delta_t$), rotation ($\delta_r$), and joint angles ($\delta_q$). We generate eight samples for each intended grasp under identical input conditions, and each sample is then sent for refinement through the quality component. For calculation purposes, $\delta_r$ and $\delta_q$ are converted into Euler angles (in degrees), and $\delta_t$ is measured in centimeters.

\begin{table*}[t]
\centering
\setlength{\tabcolsep}{5pt}
% \begin{tabular}{c|lllll|lllll}
\begin{tabular}{c|ccccc|ccccc}
\toprule

\multirow{2}{*}{} & \multicolumn{5}{c|}{\textit{\textbf{Open Set A}}} & \multicolumn{5}{c}{\textit{\textbf{Open Set B}}} \\ \cline{2-11} 
 & $FID \downarrow$ & $CD \downarrow$ & $Top1 \uparrow$ & $Q1 \uparrow$ & $Pen. \downarrow$ & $FID \downarrow$ & $CD \downarrow$ & $Top1 \uparrow$ & $Q1 \uparrow$ & $Pen. \downarrow$ \\ \midrule

\multicolumn{1}{l|}{DDIM \cite{song2020ddim}} &0.247  &4.13  &0.455  &0.0167  &0.541  &0.211  &3.66 &0.520  &0.0167  &0.542  \\ 
\multicolumn{1}{l|}{DDPM \cite{ho2020ddpm}} &0.254 &3.97  &0.461  &0.0283  &0.463  &0.204  &3.50  &0.516  &0.0128  &0.602  \\ 
\multicolumn{1}{l|}{Flow Matching} &0.242  &3.79 &0.480  &0.0240  &0.501  &0.176  &2.76 &0.538  &0.0150  &0.612   \\ 
\bottomrule
\end{tabular}
\caption{The comparison of Flow Matching with Diffusion models with DDPM and DDIM. }
\label{table:fm}
\end{table*}

\begin{table*}[t]
\centering
\begin{tabular}{ccccc}
\toprule
& AFM & GFM & DDIM & DDPM  \\ 
\midrule
Time & \(0.075 \pm 0.006\) & \(0.233 \pm 0.022\) & \(0.235 \pm 0.013\) & \(1.14 \pm 0.099\)  \\ 
\bottomrule
\end{tabular}
\caption{The inference time of Affordance Flow Matching (AFM) (first column); Grasp Flow Matching (second column); Grasp generation with DDIM and DDPM (The third and fourth columns).}
\label{table:fm_infer}
\end{table*}

\begin{table*}[t]
\centering
\begin{tabular}{c|cc|cc}
\hline
 Point IOU$\uparrow$ & \multicolumn{2}{c|}{Proposed Affordance} & \multicolumn{2}{c}{Contact Map} \\ \cline{2-5}
 & Train & Test & Train & Test \\ \hline
Set A & 0.597 & 0.508 & 0.576 & 0.359 \\ \hline
Set B & 0.602 & 0.502 & 0.624 & 0.315 \\ \hline
\end{tabular}
\caption{Generation performance of the proposed affordance and contact map.}
\label{table: affordance}
\end{table*}

\subsection{Reproduction of SOTA method}
We reproduce SOTA methods on our Open Set dataset using the same encoder structure to ensure fair comparison. Specifically, we reimplement ContactGen \cite{liu2023contactgen}, Contact2Grasp \cite{li2022contact2grasp}, GraspCAVE based on \cite{sohn2015cvae}, SceneDiffuser \cite{huang2023diffusion}, and DexGYS \cite{wei2025grasp}. To introduce language information, we use an identical CLIP language encoder. For GraspCAVE, ContactGen and Contact2Grasp, we concatenate the language feature, object feature, and latent feature to send to the decoder. For SceneDiffuser, we concatenate the language and object features as the model condition. The input language guidance of all methods are indentical as the MLLM's output for fair comparison.

\section{More Experiments}
\subsection{Effectiveness of Flow matching model}
The results in Table \ref{table:fm} and \ref{table:fm_infer} shows the performance of Our flow matching based model with the diffusion based model. It is noticed that our model surpass the diffusion based model with higher performance and faster inference speed. In addition, the introduced of the affordance flow matching only increased the time by 0.075 second compared to the baseline.

\subsection{Experiments of Affordance}

\textbf{Qualitative Experiments.} As shown in Figure \ref{fig: affcompare} and \ref{fig: graspcompare}, our generalizable-instructive affordance is more generalizable and instructive than other representations on open set. The language condition, without low level affordance condition struggles to generliza to unseen categories. The contact map is too fine-gained to generalize and the object part is too coarse to provide effective guidance. 

\noindent\textbf{Quantitative Experiments.} We conduct an experiment to prove the proposed affordance are better than contact map in generalization, as shown in Tab.~\ref{table: affordance}. The Point IOU metric is used for measure the consistency of perdition and the ground truth. The performance of contact maps drops markedly in testing, compared with affordance.

% \subsection{Qualitative Experiments of Affordance}
% As shown in Figure \ref{fig: affcompare} and \ref{fig: graspcompare}, our generalizable-instructive affordance is more generalizable and instructive than other representations on open set. The language condition, without low level affordance condition struggles to generliza to unseen categories. The contact map is too fine-gained to generalize and the object part is too coarse to provide effective guidance. 

\subsection{Qualitative Experiments of Affordance-guided Optimazation}
\begin{figure*}[!t]
\centering
\includegraphics[width=0.8\linewidth]{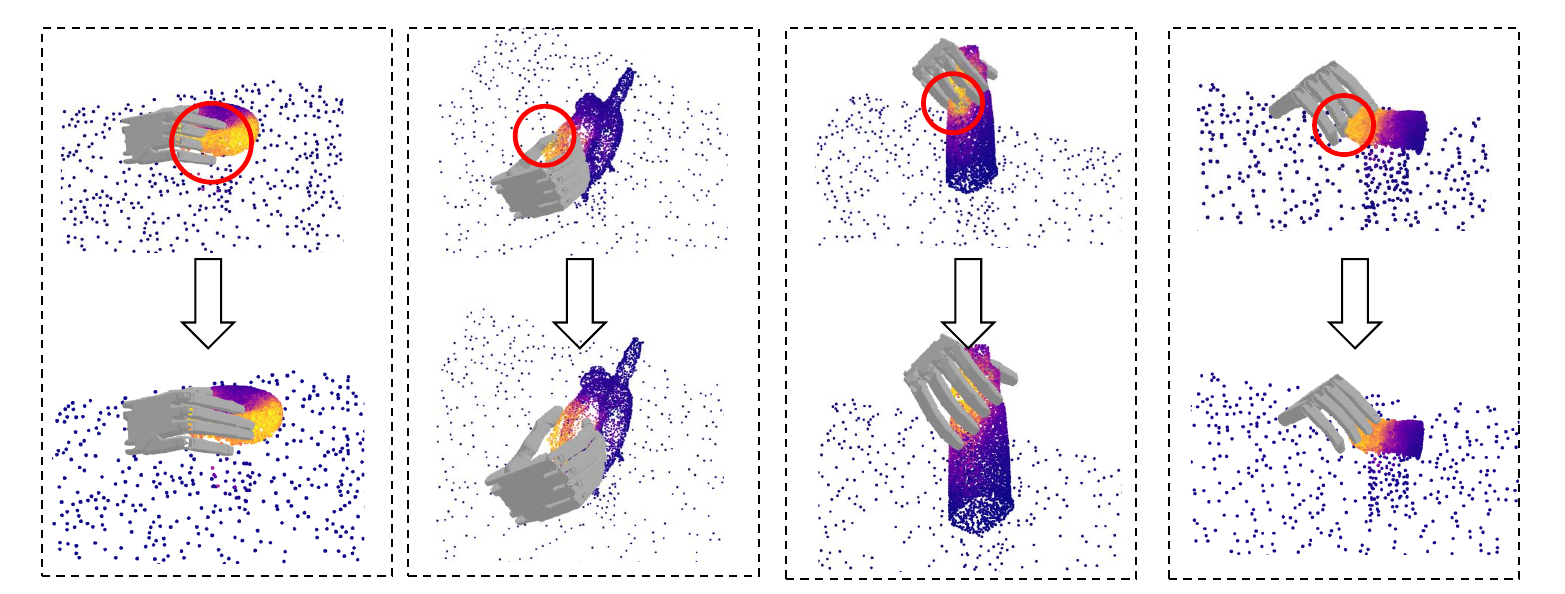}
\caption{Visualization of grasps before and after affordance-guided grasp optimization. }
\label{fig: vis_tta}
\end{figure*}

To verify the effectiveness of the affordance-guided optimazation, we offer more qualitative results. Figure~\ref{fig: vis_tta} shows the grasping situations before and after optimization. It clearly shows that the QGC can prevent the grasp from penetrating the object and keep in line with the original intention.

\subsection{Extension to Dexterous Manipulation}
\begin{figure*}[t]
\centering
\includegraphics[width=\linewidth]{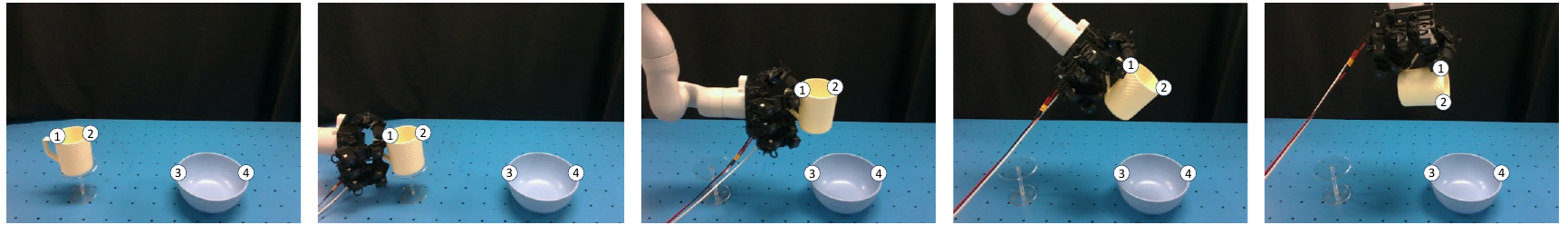}
\caption{Visualization of the application of our method to the manipulation task.}
\label{fig: real world rekep}
\end{figure*}

We extend our methods to the manipulation task by cooperating the grasp ability of our framework with task planning ability of MLLM and the perception ability of computer vision foundation model to achieve dexterous manipulation task, like pouring water, following ReKep \cite{huang2024rekep}. Specially, the MLLM first generates the relational keypoint constraints, then obtain an action solution by optimization solver. Then we employ our framework to generate a stable dexterous grasping and execute manipulation action according to the action solution. The visualization results can be found in Figure \ref{fig: real world rekep}.

\section{Real World Experiments Details}
\begin{figure*}[t]
\centering
\includegraphics[width=\linewidth]{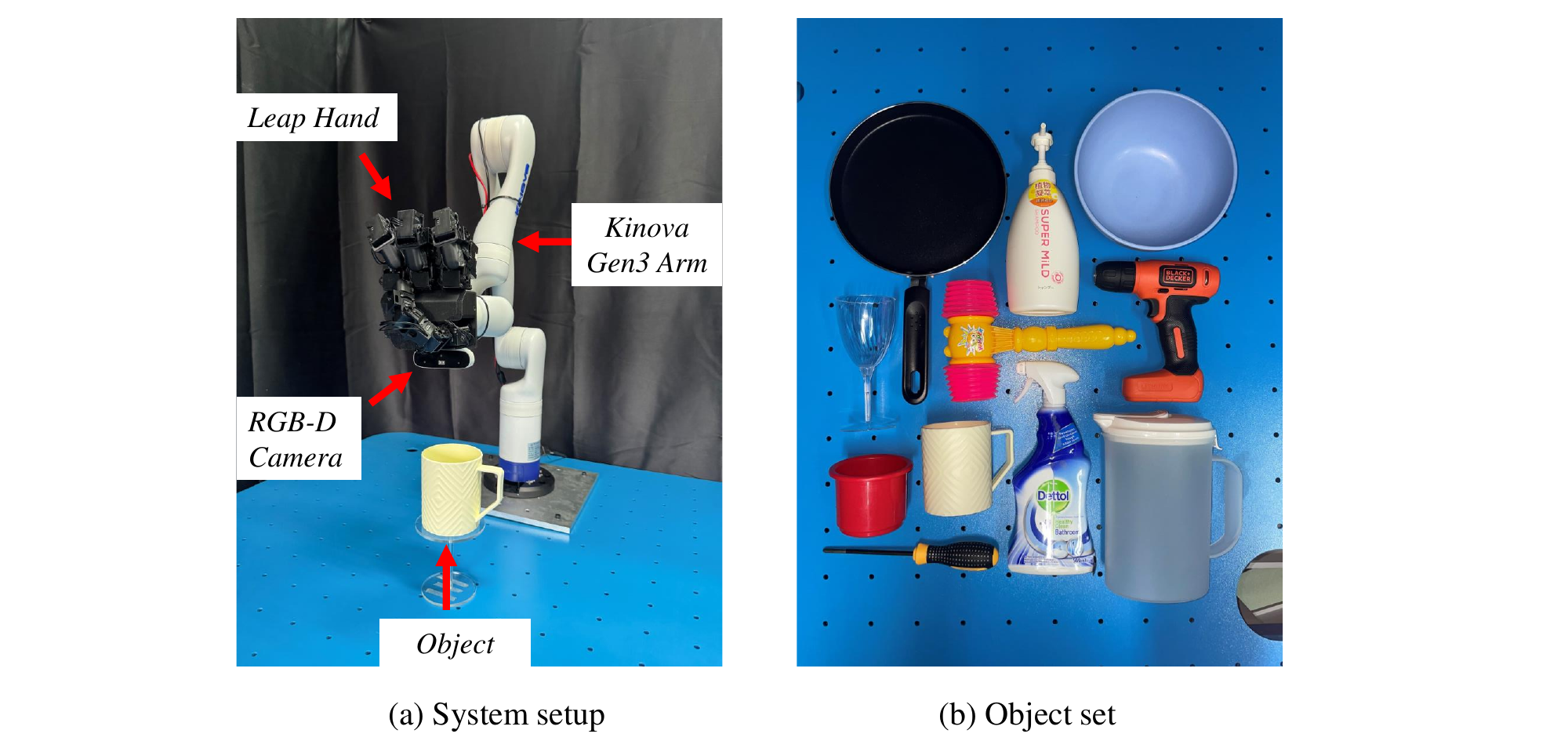}
\caption{The illustration of our real world experiment settings.}
\label{fig: real world setting}
\end{figure*}

\textbf{Experimental Environment}
Figure \ref{fig: real world setting} shows the settings of our real world experiments. The experiments are conducted on Leap hand, a Kinnova Gen3 6DOF arm and a Kinova original wrist RGB-D camera. We collect several daily life objects to evalate the open-set generalization in real world.

\begin{figure*}[t]
\centering
\includegraphics[width=\linewidth]{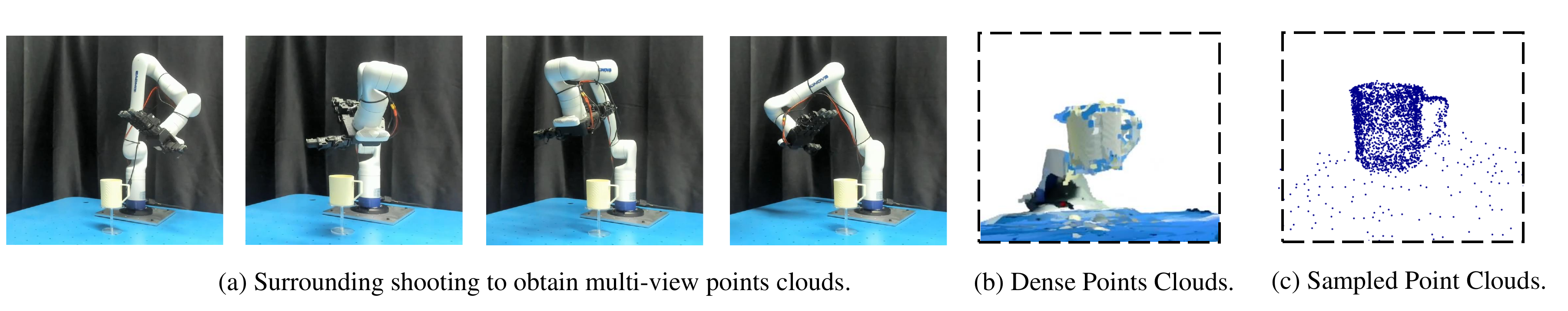}
\caption{The visualization of the pipeline to obtain scene point clouds.}
\label{fig: shotting}
\end{figure*}

\begin{figure*}[t]
\centering
\includegraphics[width=\linewidth]{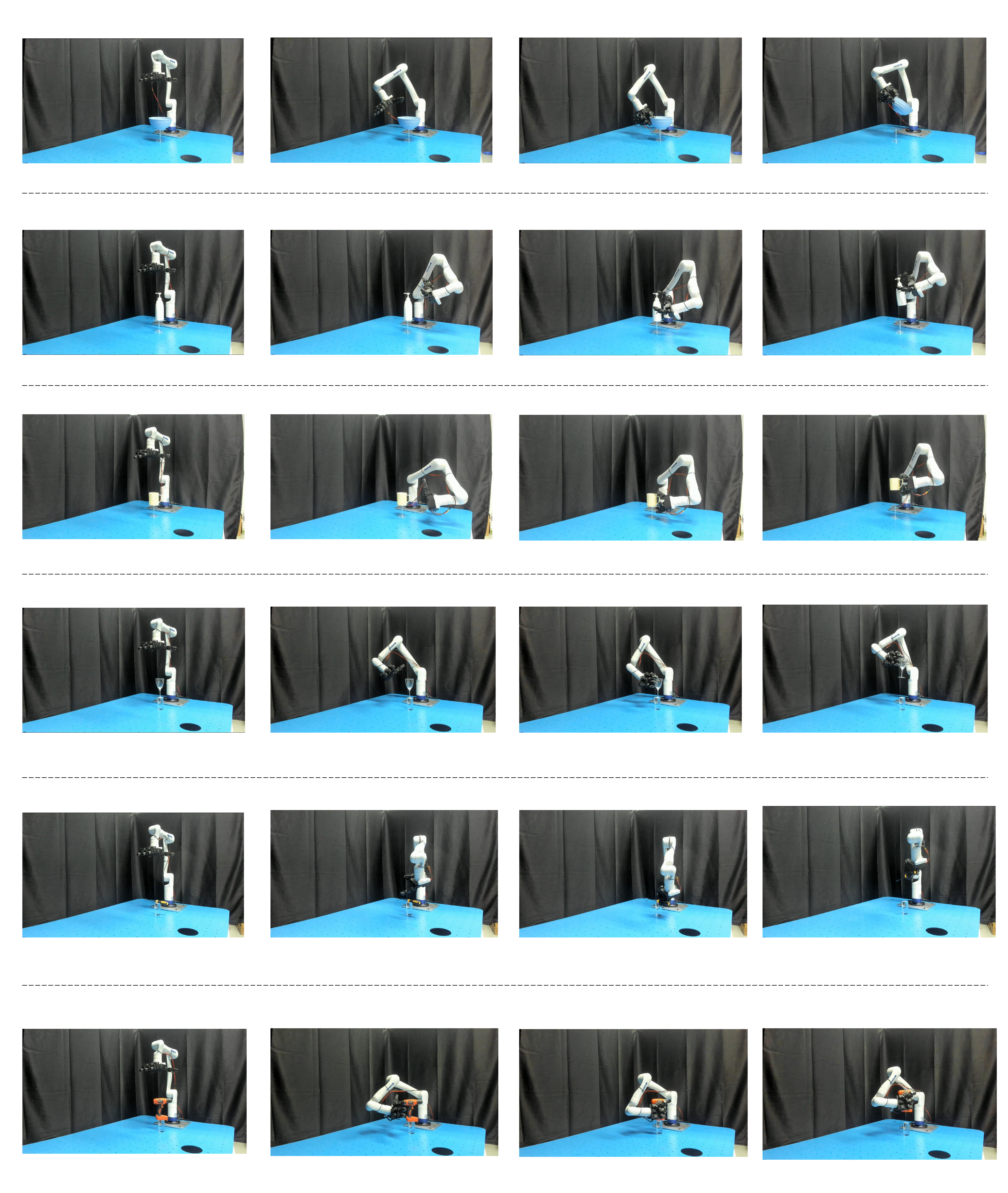}
\caption{The visualization of motion in real world experiments.}
\label{fig: realmotion}
\end{figure*}

\noindent\textbf{Experiment Pipeline}
To obtain the scene point clouds, we set the robotic arm to record partial point clouds along a specified trajectory, then fuse the local point clouds into a global point cloud and down-sample them to obtain the sampled point clouds, as shown in Figure \ref{fig: shotting}. Next, the RGB image and user commands are fed into MLLM to obtain the guidance. The guidance and scene point clouds are then fed into our framework to sequentially generate affordance and grasp poses. During the grasp execution phase, we first move the robotic arm to the vicinity of the target, and then simultaneously control both the robotic arm and the dexterous hand to perform the grasp. The visualization can be found in Figure \ref{fig: realmotion}.

% \clearpage
% \newpage
% \clearpage
% {
%     \small
%     \bibliographystyle{ieeenat_fullname}
%     \bibliography{main}
% }

\end{document}